%% file: main.tex
\documentclass[a4paper]{article}

\usepackage{INTERSPEECH2021}

\usepackage{pgfplots}

\title{How Much do Lyrics Matter? Analysing Lyrical Simplicity Preferences for Individuals At Risk of Depression}
\name{Jaidev Shriram$^1$, Sreeharsha Paruchuri$^1$, Vinoo Alluri$^1$}
\address{
  $^1$International Insitute of Information Technology, Hyderabad
  }
\email{\{jaidev.shriram, venkata.surya\}@students.iiit.ac.in, vinoo.alluri@iiit.ac.in}

\begin{document}

\maketitle

\begin{abstract}

\input{LaTeX/abstract}

\end{abstract}


\section{Introduction}

\input{LaTeX/intro}

\section{Background}
\label{Background}

\input{LaTeX/background}

\section{Methodology}
\label{Methodology}

\input{LaTeX/methodology}

\section{Statistical Testing and Results}
\label{Stats}

\input{LaTeX/stats}

\section{Discussion}
\label{Discussion}

\input{LaTeX/discussion}

\bibliographystyle{IEEEtran}

\bibliography{mybib}

\end{document}

%% file: LaTeX/abstract.tex
Music affects and in some cases reflects one's emotional state. Key to this influence is lyrics and their meaning in conjunction with the acoustic properties of the track. Recent work has focused on analysing these acoustic properties and showing that individuals prone to depression primarily consume low valence and low energy music. However, no studies yet have explored lyrical content preferences in relation to online music consumption of such individuals. In the current study, we examine lyrical simplicity, measured as the \textit{Compressibility} and Absolute Information Content of the text, associated with preferences of individuals at risk for depression. Using the six-month listening history of 541 Last.fm users, we compare lyrical simplicity trends for users grouped as being at risk (At-Risk) of depression from those that are not (No-Risk). Our findings reveal that At-Risk individuals prefer songs with greater information content (lower \textit{Compressibility}) on average, especially for songs characterised as \textit{Sad}. Furthermore, we found that At-Risk individuals also have greater variability of Absolute Information Content across their listening history. We discuss the results in light of existing socio-psychological lab-based research on music habits associated with depression and their relevance to naturally occurring online music listening behaviour.





%% file: LaTeX/intro.tex
Music plays a key role in life, often functioning as mood regulators and an expression of self \cite{schafer2013}. Due to the increasing popularity of online streaming platforms, recent studies have been able to observe such behaviour using long term listening histories. Such work has explored the relation between music preferences and personality \cite{spotifypersonality} \cite{ferwerda2015} and observed significant differences between various user groups, such as on the `Big Five Personality Traits' \cite{rocas2002}. More relevant to our work, music listening behaviour has also been found to be associated with those at risk for depression and heightened anxiety \cite{garrido2015} \cite{miranda2020}, increasingly pressing problems in modern times \cite{palgi2020}.

Prior behavioural research has shown that individuals who suffer from and are at risk for depression have a preference for sad music and display tendencies of rumination through repetitive listening, avoidance of problems through music, and gravitation towards sadness and anger \cite{saarikallio2015} \cite{garrido2015} \cite{garrido2015_2}. More recently, these listening patterns have been investigated as they occur on online music streaming platforms \cite{surana2020tag2risk} \cite{surana2020static} using listening history derived from Last.fm, albeit focusing on associated acoustic content and user-annotated tags. However, none of these studies examine lyrics, especially when lyrics are known to play a vital role in eliciting emotions \cite{Demetriou2018VocalsIM}, particularly \textit{Sadness} \cite{brattico2011}, and are associated with personal music preferences \cite{Qui2019}.

In this work, we focus on the structural properties of lyrics and the amount of information they contain. Unlike prose and other text, music is a heavily structured medium that relies on structural regularities. The typical \textit{`verse-chorus-verse-chorus-bridge-chorus}' structure for instance, features a chorus that repeats itself three times, forming a significant portion of the song. These structures and the prominence of sub-sections also vary greatly from song to song and across genres, leading to varying levels of complexity. Therefore, lyrical regularities may provide insight into individual preferences, which in turn may be a predictor for depression risk. 
 
This paper is organised as follows: Section \ref{Background} summarises prior work in this domain and provides bases for hypotheses, Section \ref{Methodology} describes the dataset used and metrics used to evaluate lyrical simplicity, Section \ref{Stats} reports the results obtained, and Section \ref{Discussion} discusses the results.

%% file: LaTeX/background.tex
\subsection{Inferring Depression Risk From Listening Habits}

Recent studies have examined the digital musical footprint of users streaming music on Last.fm to infer various characteristics of user groups prone to depression (At-Risk) and those that aren't (No-Risk). \cite{surana2020tag2risk} was the first to consider long term listening history and user-annotated tags from Last.fm, and found that At-Risk individuals listen more to music tagged as \textit{sad}, in line with \cite{garrido2015}, in addition to a preference for genres related to those, represented by tags \textit{neo-psychedelic, dream-pop} and \textit{indiepop}. 
Subsequent work \cite{surana2020static} built on this by considering emotions extracted from acoustic features over six-months and examining time-varying music consumption. This study was performed on the premise that there exist dynamic patterns of individuals suffering from or at risk for depression. Such individuals have been associated with high inertia in certain mood states in addition to heightened variability in emotions \cite{emotiondynamics} \cite{koval2016}. In line with this, \cite{surana2020static} found that At-Risk individuals show high inertia and high variability in their dynamic music listening habits. However, no study, to the best of our knowledge, has looked at lyrical regularities in a similar context.

\subsection{Lyrical Structure and Regularities}

Lyrics can be characterised in several ways, such as emotions or moods, topics, or structural regularities. Since evaluating conveyed/perceived emotions or moods may differ from emotions or moods felt, we chose to pick more objective ways to capture lyrical regularities. To this end we define lyrical simplicity, that is, the amount of repetition in a song, which in turn is represented by \textit{Compressiblity}. If a song is highly structured with several repetitive elements, we consider it to be a simpler song, with lower information content. This is similar to prior work that looks at lyrical simplicity and repetition. \cite{markus2019} investigates the repetition of lyrics across genres and artists, finding, for instance, that \textit{Dance} and \textit{Electropop} are the most repetitive genres while \textit{Death metal} and \textit{Progressive Metal} are the least repetitive. They also show that the popularity of artists highly correlates with the average repetition of their songs. \cite{morris} looked at the repetitiveness of lyrics in Billboard Hot 100 across decades, finding that repetition has increased over the past decades. Inspired by this, \cite{varnum2021} investigated the reason for the increase, theorising that certain cultural factors such as immigration, alongside greater availability of novel music contributed to simplicity.

In the current study, we explore lyrical regularities in relation to individuals at risk for depression. We hypothesise that these individuals show the following:
\begin{itemize}
    
    \item since \cite{surana2020tag2risk} and \cite{surana2020static} established that At-Risk individuals have a predilection for \textit{Sad} music and that lyrical content is essential for depicting sadness in music \cite{brattico2011}, we hypothesise that overall lower lyrical simplicity associated with music consumed by such individuals
    \item since At-Risk individuals are known to exhibit emotional inertia \cite{emotiondynamics}, we hypothesise that this group to exhibit a low variance in lyrical simplicity during a continuous period of listening history.
    \item since emotions are known to fluctuate over time for At-Risk individuals, we hypothesise that individuals belonging to this group will have a higher variability for lyrical simplicity over the entire listening history.
\end{itemize}

%% file: LaTeX/methodology.tex
\begin{figure}[t]
  \centering
  \includegraphics[width=0.8\linewidth]{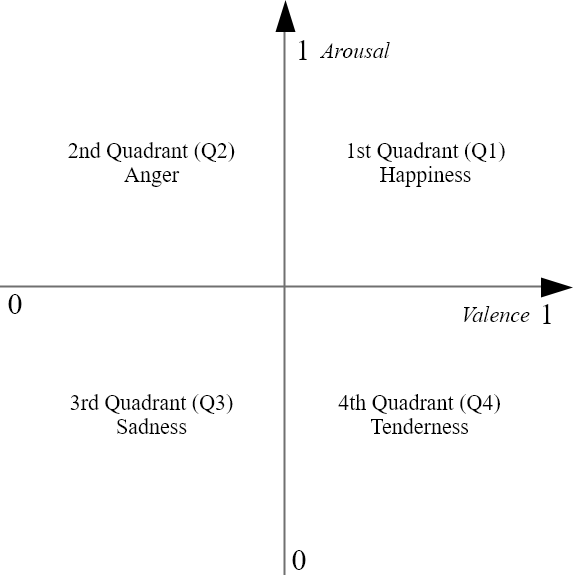}
  \caption{Two-dimensional Valence-Arousal space, representing four emotional states.}
  \label{fig:vaspace}
\end{figure}

\subsection{Dataset}
\textit{Participants: }We use the dataset in \cite{surana2020tag2risk} for our analysis. It consists of the six-month listening history of 
541 Last.fm users, ~92\% of whom fall in the 15-35 age range, with 444 males, 82 females, and 15 genderqueer in total. Each user also filled the Kessler's Psychological Distress Scale (K10) \cite{kessler_short_2002} questionnaire which measures the likelihood to have depression or anxiety on a scale from 10 to 50. Following the same procedure in \cite{surana2020tag2risk}, we consider individuals with a K10 score greater than 29 (193 individuals) as At-Risk and K10 score less than 20 (142 individuals) as No-Risk. For a more comprehensive overview of the dataset, we refer the reader to \cite{surana2020tag2risk}. 

\noindent\textit{Lyrics Extraction: }We scrape the lyrics for tracks from \textit{Genius.com} and \textit{MetroLyrics.com}, successfully obtaining 76\% of the entire listening history (3,179,462 tracks). These websites also occasionally register instrumental tracks (zero words), which comprised 4\% of the dataset. 

\subsection{Measuring Lyrical Simplicity}

Music is inherently repetitive, with repeated instances of words, phrases, and larger structures such as a chorus. Hence, the total information content of the song is smaller than the entire length of the track. We choose to represent this idea using the notion of text compression, as in \cite{morris}\cite{markus2019}. Such compression algorithms find repeated sequences of text and replace them with a shorter version, reducing the volume of data required to store the text.

We use the LZ777 compression algorithm \cite{lz777} here. The algorithm is commonly used to compress files by pointing repeated instances of text to its first instance. When considering lyrics, it is important to not look at words in isolation but rather as phrases within a longer context. Since this algorithm groups words that cluster together as one entity, akin to perceptual processes that we engage in to make contextual sense, compression related features are likely perceptually relevant as well. The final compression score is calculated as the percentage of compressible phrases. For instance, a song with just one phrase repeated many times would have a very high compression score. 

\begin{equation}
    \text{\textit{Compressibility} = }\; 1 - \frac{length(compressed\_song)}{length(song)}
\end{equation}

We assign a \textit{Compressibility} score of 1 to the songs which are determined to be purely instrumental, i.e. have no lyrics. A score of 1 also satisfies our notion of lyrical simplicity. Henceforth, we use the idea of low compressibility/complexity and high lyrical simplicity interchangeably. We also consider the `Absolute Information Content' (AIC) of the song as an additional metric. It represents the size of the compressed song. 

\begin{equation}
    \text{AIC = }\; length(compressed\_song)
\end{equation}


\subsection{Static and Dynamic Feature Analysis}

In line with \cite{surana2020static}, we delve into the static and dynamic listening habits of a user. Static features are aggregated across the listening history in an attempt to obtain an overview of each group. Dynamic features consider the time series nature of the listening history and attempt to describe the variability in lyrical patterns associated with music consumption across a time frame.

We consider the top $n$ (100, 250, 500) and \textit{all} tracks of the listening history for our \textit{static} analysis. Since different trends were observed for these thresholds in earlier work \cite{surana2020tag2risk}, we seek to explore if any such trends are visible for our features. Furthermore, previous research has shown that At-risk individuals prefer listening to music that is characterised as \textit{Sad} - low in valence and energy. We attempt to see if any inference can be made about preferences within such categories when using the entire listening history, by considering the two-dimensional \textit{Valence-Arousal} (VA) space \cite{russel1980}, as shown in Figure \ref{fig:vaspace}. Therefore, we investigate the quadrant specific \textit{Compressibility} and AIC scores of songs listened to by both groups. In order to categorise songs based on emotions, we obtained their respective Valence and Arousal features via the Spotify API, obtaining values for 83\% of the listening history. We hypothesise that At-Risk individuals have a predilection for lower lyrical simplicity and higher AIC, especially for \textit{Sad} music. Further, we conducted a non-parametric alternative to the 2$\times$2 factorial ANOVA, specifically the Permuted Wald-Type Statistic, with age and user category (At-Risk/No-Risk) as the factors to assess differences in \textit{compressibility} and AIC. We chose age groups 15-25 and 25-35 as the two levels for age as 88\% and 95\% from At-Risk and No-Risk lie within these intervals.

In order to analyse \textit{dynamic} features, the listening history for a user is split into sessions. A session is defined as a period of continuous listening, with a subsequent session occurring after an inactivity of at least two hours as described in \cite{surana2020static}\cite{gupta2018}. We investigate two metrics to analyse session based compression - intra-session variability and inter-session variability. The former metric is computed as the mean of the standard deviation of \textit{Compressibility} within sessions. Since At-Risk individuals are characterised by repetitive listening and high inertia in terms of mood-states \cite{emotiondynamics}, we hypothesise that within sessions, At-Risk individuals will have significantly lower standard deviation. inter-session variability is computed as the standard deviation of the mean \textit{Compressiblity} across user sessions. This metric captures the variance in information level of lyrics a person listens to across sessions.  Since At-Risk individuals are characterised, as mentioned above, by highly inert states, which also shift and fluctuate over time, we hypothesise that inter-session variability would be higher for them.

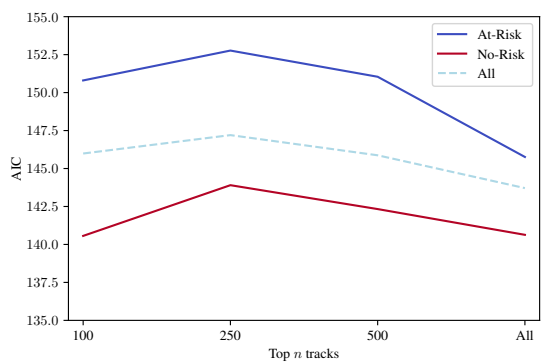
\begin{figure}[bp!]
  \centering
  \resizebox{\linewidth}{!}{\input{LaTeX/images/Infoplot.pgf}}
  \caption{Median AIC values for the At-Risk and No-Risk groups across the top $n (100, 250, 500)$ and entire listening history (All).}
  \label{fig:info_all}
\end{figure}

%% file: LaTeX/images/Infoplot.pgf
\begingroup%
\makeatletter%
\begin{pgfpicture}%
\pgfpathrectangle{\pgfpointorigin}{\pgfqpoint{6.000000in}{4.000000in}}%
\pgfusepath{use as bounding box, clip}%
\begin{pgfscope}%
\pgfsetbuttcap%
\pgfsetmiterjoin%
\pgfsetlinewidth{0.000000pt}%
\definecolor{currentstroke}{rgb}{1.000000,1.000000,1.000000}%
\pgfsetstrokecolor{currentstroke}%
\pgfsetstrokeopacity{0.000000}%
\pgfsetdash{}{0pt}%
\pgfpathmoveto{\pgfqpoint{0.000000in}{0.000000in}}%
\pgfpathlineto{\pgfqpoint{6.000000in}{0.000000in}}%
\pgfpathlineto{\pgfqpoint{6.000000in}{4.000000in}}%
\pgfpathlineto{\pgfqpoint{0.000000in}{4.000000in}}%
\pgfpathclose%
\pgfusepath{}%
\end{pgfscope}%
\begin{pgfscope}%
\pgfsetbuttcap%
\pgfsetmiterjoin%
\definecolor{currentfill}{rgb}{1.000000,1.000000,1.000000}%
\pgfsetfillcolor{currentfill}%
\pgfsetlinewidth{0.000000pt}%
\definecolor{currentstroke}{rgb}{0.000000,0.000000,0.000000}%
\pgfsetstrokecolor{currentstroke}%
\pgfsetstrokeopacity{0.000000}%
\pgfsetdash{}{0pt}%
\pgfpathmoveto{\pgfqpoint{0.750000in}{0.500000in}}%
\pgfpathlineto{\pgfqpoint{5.400000in}{0.500000in}}%
\pgfpathlineto{\pgfqpoint{5.400000in}{3.520000in}}%
\pgfpathlineto{\pgfqpoint{0.750000in}{3.520000in}}%
\pgfpathclose%
\pgfusepath{fill}%
\end{pgfscope}%
\begin{pgfscope}%
\pgfsetbuttcap%
\pgfsetroundjoin%
\definecolor{currentfill}{rgb}{0.000000,0.000000,0.000000}%
\pgfsetfillcolor{currentfill}%
\pgfsetlinewidth{0.803000pt}%
\definecolor{currentstroke}{rgb}{0.000000,0.000000,0.000000}%
\pgfsetstrokecolor{currentstroke}%
\pgfsetdash{}{0pt}%
\pgfsys@defobject{currentmarker}{\pgfqpoint{0.000000in}{-0.048611in}}{\pgfqpoint{0.000000in}{0.000000in}}{%
\pgfpathmoveto{\pgfqpoint{0.000000in}{0.000000in}}%
\pgfpathlineto{\pgfqpoint{0.000000in}{-0.048611in}}%
\pgfusepath{stroke,fill}%
}%
\begin{pgfscope}%
\pgfsys@transformshift{0.895313in}{0.500000in}%
\pgfsys@useobject{currentmarker}{}%
\end{pgfscope}%
\end{pgfscope}%
\begin{pgfscope}%
\definecolor{textcolor}{rgb}{0.000000,0.000000,0.000000}%
\pgfsetstrokecolor{textcolor}%
\pgfsetfillcolor{textcolor}%
\pgftext[x=0.895313in,y=0.402778in,,top]{\color{textcolor}\rmfamily\fontsize{10.000000}{12.000000}\selectfont 100}%
\end{pgfscope}%
\begin{pgfscope}%
\pgfsetbuttcap%
\pgfsetroundjoin%
\definecolor{currentfill}{rgb}{0.000000,0.000000,0.000000}%
\pgfsetfillcolor{currentfill}%
\pgfsetlinewidth{0.803000pt}%
\definecolor{currentstroke}{rgb}{0.000000,0.000000,0.000000}%
\pgfsetstrokecolor{currentstroke}%
\pgfsetdash{}{0pt}%
\pgfsys@defobject{currentmarker}{\pgfqpoint{0.000000in}{-0.048611in}}{\pgfqpoint{0.000000in}{0.000000in}}{%
\pgfpathmoveto{\pgfqpoint{0.000000in}{0.000000in}}%
\pgfpathlineto{\pgfqpoint{0.000000in}{-0.048611in}}%
\pgfusepath{stroke,fill}%
}%
\begin{pgfscope}%
\pgfsys@transformshift{2.348438in}{0.500000in}%
\pgfsys@useobject{currentmarker}{}%
\end{pgfscope}%
\end{pgfscope}%
\begin{pgfscope}%
\definecolor{textcolor}{rgb}{0.000000,0.000000,0.000000}%
\pgfsetstrokecolor{textcolor}%
\pgfsetfillcolor{textcolor}%
\pgftext[x=2.348438in,y=0.402778in,,top]{\color{textcolor}\rmfamily\fontsize{10.000000}{12.000000}\selectfont 250}%
\end{pgfscope}%
\begin{pgfscope}%
\pgfsetbuttcap%
\pgfsetroundjoin%
\definecolor{currentfill}{rgb}{0.000000,0.000000,0.000000}%
\pgfsetfillcolor{currentfill}%
\pgfsetlinewidth{0.803000pt}%
\definecolor{currentstroke}{rgb}{0.000000,0.000000,0.000000}%
\pgfsetstrokecolor{currentstroke}%
\pgfsetdash{}{0pt}%
\pgfsys@defobject{currentmarker}{\pgfqpoint{0.000000in}{-0.048611in}}{\pgfqpoint{0.000000in}{0.000000in}}{%
\pgfpathmoveto{\pgfqpoint{0.000000in}{0.000000in}}%
\pgfpathlineto{\pgfqpoint{0.000000in}{-0.048611in}}%
\pgfusepath{stroke,fill}%
}%
\begin{pgfscope}%
\pgfsys@transformshift{3.801563in}{0.500000in}%
\pgfsys@useobject{currentmarker}{}%
\end{pgfscope}%
\end{pgfscope}%
\begin{pgfscope}%
\definecolor{textcolor}{rgb}{0.000000,0.000000,0.000000}%
\pgfsetstrokecolor{textcolor}%
\pgfsetfillcolor{textcolor}%
\pgftext[x=3.801563in,y=0.402778in,,top]{\color{textcolor}\rmfamily\fontsize{10.000000}{12.000000}\selectfont 500}%
\end{pgfscope}%
\begin{pgfscope}%
\pgfsetbuttcap%
\pgfsetroundjoin%
\definecolor{currentfill}{rgb}{0.000000,0.000000,0.000000}%
\pgfsetfillcolor{currentfill}%
\pgfsetlinewidth{0.803000pt}%
\definecolor{currentstroke}{rgb}{0.000000,0.000000,0.000000}%
\pgfsetstrokecolor{currentstroke}%
\pgfsetdash{}{0pt}%
\pgfsys@defobject{currentmarker}{\pgfqpoint{0.000000in}{-0.048611in}}{\pgfqpoint{0.000000in}{0.000000in}}{%
\pgfpathmoveto{\pgfqpoint{0.000000in}{0.000000in}}%
\pgfpathlineto{\pgfqpoint{0.000000in}{-0.048611in}}%
\pgfusepath{stroke,fill}%
}%
\begin{pgfscope}%
\pgfsys@transformshift{5.254688in}{0.500000in}%
\pgfsys@useobject{currentmarker}{}%
\end{pgfscope}%
\end{pgfscope}%
\begin{pgfscope}%
\definecolor{textcolor}{rgb}{0.000000,0.000000,0.000000}%
\pgfsetstrokecolor{textcolor}%
\pgfsetfillcolor{textcolor}%
\pgftext[x=5.254688in,y=0.402778in,,top]{\color{textcolor}\rmfamily\fontsize{10.000000}{12.000000}\selectfont All}%
\end{pgfscope}%
\begin{pgfscope}%
\definecolor{textcolor}{rgb}{0.000000,0.000000,0.000000}%
\pgfsetstrokecolor{textcolor}%
\pgfsetfillcolor{textcolor}%
\pgftext[x=3.075000in,y=0.223766in,,top]{\color{textcolor}\rmfamily\fontsize{10.000000}{12.000000}\selectfont Top \(\displaystyle n\) tracks}%
\end{pgfscope}%
\begin{pgfscope}%
\pgfsetbuttcap%
\pgfsetroundjoin%
\definecolor{currentfill}{rgb}{0.000000,0.000000,0.000000}%
\pgfsetfillcolor{currentfill}%
\pgfsetlinewidth{0.803000pt}%
\definecolor{currentstroke}{rgb}{0.000000,0.000000,0.000000}%
\pgfsetstrokecolor{currentstroke}%
\pgfsetdash{}{0pt}%
\pgfsys@defobject{currentmarker}{\pgfqpoint{-0.048611in}{0.000000in}}{\pgfqpoint{-0.000000in}{0.000000in}}{%
\pgfpathmoveto{\pgfqpoint{-0.000000in}{0.000000in}}%
\pgfpathlineto{\pgfqpoint{-0.048611in}{0.000000in}}%
\pgfusepath{stroke,fill}%
}%
\begin{pgfscope}%
\pgfsys@transformshift{0.750000in}{0.500000in}%
\pgfsys@useobject{currentmarker}{}%
\end{pgfscope}%
\end{pgfscope}%
\begin{pgfscope}%
\definecolor{textcolor}{rgb}{0.000000,0.000000,0.000000}%
\pgfsetstrokecolor{textcolor}%
\pgfsetfillcolor{textcolor}%
\pgftext[x=0.336419in, y=0.451775in, left, base]{\color{textcolor}\rmfamily\fontsize{10.000000}{12.000000}\selectfont \(\displaystyle {135.0}\)}%
\end{pgfscope}%
\begin{pgfscope}%
\pgfsetbuttcap%
\pgfsetroundjoin%
\definecolor{currentfill}{rgb}{0.000000,0.000000,0.000000}%
\pgfsetfillcolor{currentfill}%
\pgfsetlinewidth{0.803000pt}%
\definecolor{currentstroke}{rgb}{0.000000,0.000000,0.000000}%
\pgfsetstrokecolor{currentstroke}%
\pgfsetdash{}{0pt}%
\pgfsys@defobject{currentmarker}{\pgfqpoint{-0.048611in}{0.000000in}}{\pgfqpoint{-0.000000in}{0.000000in}}{%
\pgfpathmoveto{\pgfqpoint{-0.000000in}{0.000000in}}%
\pgfpathlineto{\pgfqpoint{-0.048611in}{0.000000in}}%
\pgfusepath{stroke,fill}%
}%
\begin{pgfscope}%
\pgfsys@transformshift{0.750000in}{0.877500in}%
\pgfsys@useobject{currentmarker}{}%
\end{pgfscope}%
\end{pgfscope}%
\begin{pgfscope}%
\definecolor{textcolor}{rgb}{0.000000,0.000000,0.000000}%
\pgfsetstrokecolor{textcolor}%
\pgfsetfillcolor{textcolor}%
\pgftext[x=0.336419in, y=0.829275in, left, base]{\color{textcolor}\rmfamily\fontsize{10.000000}{12.000000}\selectfont \(\displaystyle {137.5}\)}%
\end{pgfscope}%
\begin{pgfscope}%
\pgfsetbuttcap%
\pgfsetroundjoin%
\definecolor{currentfill}{rgb}{0.000000,0.000000,0.000000}%
\pgfsetfillcolor{currentfill}%
\pgfsetlinewidth{0.803000pt}%
\definecolor{currentstroke}{rgb}{0.000000,0.000000,0.000000}%
\pgfsetstrokecolor{currentstroke}%
\pgfsetdash{}{0pt}%
\pgfsys@defobject{currentmarker}{\pgfqpoint{-0.048611in}{0.000000in}}{\pgfqpoint{-0.000000in}{0.000000in}}{%
\pgfpathmoveto{\pgfqpoint{-0.000000in}{0.000000in}}%
\pgfpathlineto{\pgfqpoint{-0.048611in}{0.000000in}}%
\pgfusepath{stroke,fill}%
}%
\begin{pgfscope}%
\pgfsys@transformshift{0.750000in}{1.255000in}%
\pgfsys@useobject{currentmarker}{}%
\end{pgfscope}%
\end{pgfscope}%
\begin{pgfscope}%
\definecolor{textcolor}{rgb}{0.000000,0.000000,0.000000}%
\pgfsetstrokecolor{textcolor}%
\pgfsetfillcolor{textcolor}%
\pgftext[x=0.336419in, y=1.206775in, left, base]{\color{textcolor}\rmfamily\fontsize{10.000000}{12.000000}\selectfont \(\displaystyle {140.0}\)}%
\end{pgfscope}%
\begin{pgfscope}%
\pgfsetbuttcap%
\pgfsetroundjoin%
\definecolor{currentfill}{rgb}{0.000000,0.000000,0.000000}%
\pgfsetfillcolor{currentfill}%
\pgfsetlinewidth{0.803000pt}%
\definecolor{currentstroke}{rgb}{0.000000,0.000000,0.000000}%
\pgfsetstrokecolor{currentstroke}%
\pgfsetdash{}{0pt}%
\pgfsys@defobject{currentmarker}{\pgfqpoint{-0.048611in}{0.000000in}}{\pgfqpoint{-0.000000in}{0.000000in}}{%
\pgfpathmoveto{\pgfqpoint{-0.000000in}{0.000000in}}%
\pgfpathlineto{\pgfqpoint{-0.048611in}{0.000000in}}%
\pgfusepath{stroke,fill}%
}%
\begin{pgfscope}%
\pgfsys@transformshift{0.750000in}{1.632500in}%
\pgfsys@useobject{currentmarker}{}%
\end{pgfscope}%
\end{pgfscope}%
\begin{pgfscope}%
\definecolor{textcolor}{rgb}{0.000000,0.000000,0.000000}%
\pgfsetstrokecolor{textcolor}%
\pgfsetfillcolor{textcolor}%
\pgftext[x=0.336419in, y=1.584275in, left, base]{\color{textcolor}\rmfamily\fontsize{10.000000}{12.000000}\selectfont \(\displaystyle {142.5}\)}%
\end{pgfscope}%
\begin{pgfscope}%
\pgfsetbuttcap%
\pgfsetroundjoin%
\definecolor{currentfill}{rgb}{0.000000,0.000000,0.000000}%
\pgfsetfillcolor{currentfill}%
\pgfsetlinewidth{0.803000pt}%
\definecolor{currentstroke}{rgb}{0.000000,0.000000,0.000000}%
\pgfsetstrokecolor{currentstroke}%
\pgfsetdash{}{0pt}%
\pgfsys@defobject{currentmarker}{\pgfqpoint{-0.048611in}{0.000000in}}{\pgfqpoint{-0.000000in}{0.000000in}}{%
\pgfpathmoveto{\pgfqpoint{-0.000000in}{0.000000in}}%
\pgfpathlineto{\pgfqpoint{-0.048611in}{0.000000in}}%
\pgfusepath{stroke,fill}%
}%
\begin{pgfscope}%
\pgfsys@transformshift{0.750000in}{2.010000in}%
\pgfsys@useobject{currentmarker}{}%
\end{pgfscope}%
\end{pgfscope}%
\begin{pgfscope}%
\definecolor{textcolor}{rgb}{0.000000,0.000000,0.000000}%
\pgfsetstrokecolor{textcolor}%
\pgfsetfillcolor{textcolor}%
\pgftext[x=0.336419in, y=1.961775in, left, base]{\color{textcolor}\rmfamily\fontsize{10.000000}{12.000000}\selectfont \(\displaystyle {145.0}\)}%
\end{pgfscope}%
\begin{pgfscope}%
\pgfsetbuttcap%
\pgfsetroundjoin%
\definecolor{currentfill}{rgb}{0.000000,0.000000,0.000000}%
\pgfsetfillcolor{currentfill}%
\pgfsetlinewidth{0.803000pt}%
\definecolor{currentstroke}{rgb}{0.000000,0.000000,0.000000}%
\pgfsetstrokecolor{currentstroke}%
\pgfsetdash{}{0pt}%
\pgfsys@defobject{currentmarker}{\pgfqpoint{-0.048611in}{0.000000in}}{\pgfqpoint{-0.000000in}{0.000000in}}{%
\pgfpathmoveto{\pgfqpoint{-0.000000in}{0.000000in}}%
\pgfpathlineto{\pgfqpoint{-0.048611in}{0.000000in}}%
\pgfusepath{stroke,fill}%
}%
\begin{pgfscope}%
\pgfsys@transformshift{0.750000in}{2.387500in}%
\pgfsys@useobject{currentmarker}{}%
\end{pgfscope}%
\end{pgfscope}%
\begin{pgfscope}%
\definecolor{textcolor}{rgb}{0.000000,0.000000,0.000000}%
\pgfsetstrokecolor{textcolor}%
\pgfsetfillcolor{textcolor}%
\pgftext[x=0.336419in, y=2.339275in, left, base]{\color{textcolor}\rmfamily\fontsize{10.000000}{12.000000}\selectfont \(\displaystyle {147.5}\)}%
\end{pgfscope}%
\begin{pgfscope}%
\pgfsetbuttcap%
\pgfsetroundjoin%
\definecolor{currentfill}{rgb}{0.000000,0.000000,0.000000}%
\pgfsetfillcolor{currentfill}%
\pgfsetlinewidth{0.803000pt}%
\definecolor{currentstroke}{rgb}{0.000000,0.000000,0.000000}%
\pgfsetstrokecolor{currentstroke}%
\pgfsetdash{}{0pt}%
\pgfsys@defobject{currentmarker}{\pgfqpoint{-0.048611in}{0.000000in}}{\pgfqpoint{-0.000000in}{0.000000in}}{%
\pgfpathmoveto{\pgfqpoint{-0.000000in}{0.000000in}}%
\pgfpathlineto{\pgfqpoint{-0.048611in}{0.000000in}}%
\pgfusepath{stroke,fill}%
}%
\begin{pgfscope}%
\pgfsys@transformshift{0.750000in}{2.765000in}%
\pgfsys@useobject{currentmarker}{}%
\end{pgfscope}%
\end{pgfscope}%
\begin{pgfscope}%
\definecolor{textcolor}{rgb}{0.000000,0.000000,0.000000}%
\pgfsetstrokecolor{textcolor}%
\pgfsetfillcolor{textcolor}%
\pgftext[x=0.336419in, y=2.716775in, left, base]{\color{textcolor}\rmfamily\fontsize{10.000000}{12.000000}\selectfont \(\displaystyle {150.0}\)}%
\end{pgfscope}%
\begin{pgfscope}%
\pgfsetbuttcap%
\pgfsetroundjoin%
\definecolor{currentfill}{rgb}{0.000000,0.000000,0.000000}%
\pgfsetfillcolor{currentfill}%
\pgfsetlinewidth{0.803000pt}%
\definecolor{currentstroke}{rgb}{0.000000,0.000000,0.000000}%
\pgfsetstrokecolor{currentstroke}%
\pgfsetdash{}{0pt}%
\pgfsys@defobject{currentmarker}{\pgfqpoint{-0.048611in}{0.000000in}}{\pgfqpoint{-0.000000in}{0.000000in}}{%
\pgfpathmoveto{\pgfqpoint{-0.000000in}{0.000000in}}%
\pgfpathlineto{\pgfqpoint{-0.048611in}{0.000000in}}%
\pgfusepath{stroke,fill}%
}%
\begin{pgfscope}%
\pgfsys@transformshift{0.750000in}{3.142500in}%
\pgfsys@useobject{currentmarker}{}%
\end{pgfscope}%
\end{pgfscope}%
\begin{pgfscope}%
\definecolor{textcolor}{rgb}{0.000000,0.000000,0.000000}%
\pgfsetstrokecolor{textcolor}%
\pgfsetfillcolor{textcolor}%
\pgftext[x=0.336419in, y=3.094275in, left, base]{\color{textcolor}\rmfamily\fontsize{10.000000}{12.000000}\selectfont \(\displaystyle {152.5}\)}%
\end{pgfscope}%
\begin{pgfscope}%
\pgfsetbuttcap%
\pgfsetroundjoin%
\definecolor{currentfill}{rgb}{0.000000,0.000000,0.000000}%
\pgfsetfillcolor{currentfill}%
\pgfsetlinewidth{0.803000pt}%
\definecolor{currentstroke}{rgb}{0.000000,0.000000,0.000000}%
\pgfsetstrokecolor{currentstroke}%
\pgfsetdash{}{0pt}%
\pgfsys@defobject{currentmarker}{\pgfqpoint{-0.048611in}{0.000000in}}{\pgfqpoint{-0.000000in}{0.000000in}}{%
\pgfpathmoveto{\pgfqpoint{-0.000000in}{0.000000in}}%
\pgfpathlineto{\pgfqpoint{-0.048611in}{0.000000in}}%
\pgfusepath{stroke,fill}%
}%
\begin{pgfscope}%
\pgfsys@transformshift{0.750000in}{3.520000in}%
\pgfsys@useobject{currentmarker}{}%
\end{pgfscope}%
\end{pgfscope}%
\begin{pgfscope}%
\definecolor{textcolor}{rgb}{0.000000,0.000000,0.000000}%
\pgfsetstrokecolor{textcolor}%
\pgfsetfillcolor{textcolor}%
\pgftext[x=0.336419in, y=3.471775in, left, base]{\color{textcolor}\rmfamily\fontsize{10.000000}{12.000000}\selectfont \(\displaystyle {155.0}\)}%
\end{pgfscope}%
\begin{pgfscope}%
\definecolor{textcolor}{rgb}{0.000000,0.000000,0.000000}%
\pgfsetstrokecolor{textcolor}%
\pgfsetfillcolor{textcolor}%
\pgftext[x=0.280863in,y=2.010000in,,bottom,rotate=90.000000]{\color{textcolor}\rmfamily\fontsize{10.000000}{12.000000}\selectfont AIC}%
\end{pgfscope}%
\begin{pgfscope}%
\pgfpathrectangle{\pgfqpoint{0.750000in}{0.500000in}}{\pgfqpoint{4.650000in}{3.020000in}}%
\pgfusepath{clip}%
\pgfsetrectcap%
\pgfsetroundjoin%
\pgfsetlinewidth{1.505625pt}%
\definecolor{currentstroke}{rgb}{0.229806,0.298718,0.753683}%
\pgfsetstrokecolor{currentstroke}%
\pgfsetdash{}{0pt}%
\pgfpathmoveto{\pgfqpoint{0.895313in}{2.885908in}}%
\pgfpathlineto{\pgfqpoint{2.348438in}{3.182976in}}%
\pgfpathlineto{\pgfqpoint{3.801563in}{2.922528in}}%
\pgfpathlineto{\pgfqpoint{5.254688in}{2.124954in}}%
\pgfusepath{stroke}%
\end{pgfscope}%
\begin{pgfscope}%
\pgfpathrectangle{\pgfqpoint{0.750000in}{0.500000in}}{\pgfqpoint{4.650000in}{3.020000in}}%
\pgfusepath{clip}%
\pgfsetrectcap%
\pgfsetroundjoin%
\pgfsetlinewidth{1.505625pt}%
\definecolor{currentstroke}{rgb}{0.705673,0.015556,0.150233}%
\pgfsetstrokecolor{currentstroke}%
\pgfsetdash{}{0pt}%
\pgfpathmoveto{\pgfqpoint{0.895313in}{1.340046in}}%
\pgfpathlineto{\pgfqpoint{2.348438in}{1.844328in}}%
\pgfpathlineto{\pgfqpoint{3.801563in}{1.606736in}}%
\pgfpathlineto{\pgfqpoint{5.254688in}{1.350678in}}%
\pgfusepath{stroke}%
\end{pgfscope}%
\begin{pgfscope}%
\pgfpathrectangle{\pgfqpoint{0.750000in}{0.500000in}}{\pgfqpoint{4.650000in}{3.020000in}}%
\pgfusepath{clip}%
\pgfsetbuttcap%
\pgfsetroundjoin%
\pgfsetlinewidth{1.505625pt}%
\definecolor{currentstroke}{rgb}{0.678431,0.847059,0.901961}%
\pgfsetstrokecolor{currentstroke}%
\pgfsetdash{{5.550000pt}{2.400000pt}}{0.000000pt}%
\pgfpathmoveto{\pgfqpoint{0.895313in}{2.159224in}}%
\pgfpathlineto{\pgfqpoint{2.348438in}{2.342200in}}%
\pgfpathlineto{\pgfqpoint{3.801563in}{2.141533in}}%
\pgfpathlineto{\pgfqpoint{5.254688in}{1.815215in}}%
\pgfusepath{stroke}%
\end{pgfscope}%
\begin{pgfscope}%
\pgfsetrectcap%
\pgfsetmiterjoin%
\pgfsetlinewidth{0.803000pt}%
\definecolor{currentstroke}{rgb}{0.000000,0.000000,0.000000}%
\pgfsetstrokecolor{currentstroke}%
\pgfsetdash{}{0pt}%
\pgfpathmoveto{\pgfqpoint{0.750000in}{0.500000in}}%
\pgfpathlineto{\pgfqpoint{0.750000in}{3.520000in}}%
\pgfusepath{stroke}%
\end{pgfscope}%
\begin{pgfscope}%
\pgfsetrectcap%
\pgfsetmiterjoin%
\pgfsetlinewidth{0.803000pt}%
\definecolor{currentstroke}{rgb}{0.000000,0.000000,0.000000}%
\pgfsetstrokecolor{currentstroke}%
\pgfsetdash{}{0pt}%
\pgfpathmoveto{\pgfqpoint{5.400000in}{0.500000in}}%
\pgfpathlineto{\pgfqpoint{5.400000in}{3.520000in}}%
\pgfusepath{stroke}%
\end{pgfscope}%
\begin{pgfscope}%
\pgfsetrectcap%
\pgfsetmiterjoin%
\pgfsetlinewidth{0.803000pt}%
\definecolor{currentstroke}{rgb}{0.000000,0.000000,0.000000}%
\pgfsetstrokecolor{currentstroke}%
\pgfsetdash{}{0pt}%
\pgfpathmoveto{\pgfqpoint{0.750000in}{0.500000in}}%
\pgfpathlineto{\pgfqpoint{5.400000in}{0.500000in}}%
\pgfusepath{stroke}%
\end{pgfscope}%
\begin{pgfscope}%
\pgfsetrectcap%
\pgfsetmiterjoin%
\pgfsetlinewidth{0.803000pt}%
\definecolor{currentstroke}{rgb}{0.000000,0.000000,0.000000}%
\pgfsetstrokecolor{currentstroke}%
\pgfsetdash{}{0pt}%
\pgfpathmoveto{\pgfqpoint{0.750000in}{3.520000in}}%
\pgfpathlineto{\pgfqpoint{5.400000in}{3.520000in}}%
\pgfusepath{stroke}%
\end{pgfscope}%
\begin{pgfscope}%
\pgfsetbuttcap%
\pgfsetmiterjoin%
\definecolor{currentfill}{rgb}{1.000000,1.000000,1.000000}%
\pgfsetfillcolor{currentfill}%
\pgfsetfillopacity{0.800000}%
\pgfsetlinewidth{1.003750pt}%
\definecolor{currentstroke}{rgb}{0.800000,0.800000,0.800000}%
\pgfsetstrokecolor{currentstroke}%
\pgfsetstrokeopacity{0.800000}%
\pgfsetdash{}{0pt}%
\pgfpathmoveto{\pgfqpoint{4.369521in}{2.827871in}}%
\pgfpathlineto{\pgfqpoint{5.302778in}{2.827871in}}%
\pgfpathquadraticcurveto{\pgfqpoint{5.330556in}{2.827871in}}{\pgfqpoint{5.330556in}{2.855648in}}%
\pgfpathlineto{\pgfqpoint{5.330556in}{3.422778in}}%
\pgfpathquadraticcurveto{\pgfqpoint{5.330556in}{3.450556in}}{\pgfqpoint{5.302778in}{3.450556in}}%
\pgfpathlineto{\pgfqpoint{4.369521in}{3.450556in}}%
\pgfpathquadraticcurveto{\pgfqpoint{4.341743in}{3.450556in}}{\pgfqpoint{4.341743in}{3.422778in}}%
\pgfpathlineto{\pgfqpoint{4.341743in}{2.855648in}}%
\pgfpathquadraticcurveto{\pgfqpoint{4.341743in}{2.827871in}}{\pgfqpoint{4.369521in}{2.827871in}}%
\pgfpathclose%
\pgfusepath{stroke,fill}%
\end{pgfscope}%
\begin{pgfscope}%
\pgfsetrectcap%
\pgfsetroundjoin%
\pgfsetlinewidth{1.505625pt}%
\definecolor{currentstroke}{rgb}{0.229806,0.298718,0.753683}%
\pgfsetstrokecolor{currentstroke}%
\pgfsetdash{}{0pt}%
\pgfpathmoveto{\pgfqpoint{4.397299in}{3.346389in}}%
\pgfpathlineto{\pgfqpoint{4.675076in}{3.346389in}}%
\pgfusepath{stroke}%
\end{pgfscope}%
\begin{pgfscope}%
\definecolor{textcolor}{rgb}{0.000000,0.000000,0.000000}%
\pgfsetstrokecolor{textcolor}%
\pgfsetfillcolor{textcolor}%
\pgftext[x=4.786188in,y=3.297778in,left,base]{\color{textcolor}\rmfamily\fontsize{10.000000}{12.000000}\selectfont At-Risk}%
\end{pgfscope}%
\begin{pgfscope}%
\pgfsetrectcap%
\pgfsetroundjoin%
\pgfsetlinewidth{1.505625pt}%
\definecolor{currentstroke}{rgb}{0.705673,0.015556,0.150233}%
\pgfsetstrokecolor{currentstroke}%
\pgfsetdash{}{0pt}%
\pgfpathmoveto{\pgfqpoint{4.397299in}{3.152716in}}%
\pgfpathlineto{\pgfqpoint{4.675076in}{3.152716in}}%
\pgfusepath{stroke}%
\end{pgfscope}%
\begin{pgfscope}%
\definecolor{textcolor}{rgb}{0.000000,0.000000,0.000000}%
\pgfsetstrokecolor{textcolor}%
\pgfsetfillcolor{textcolor}%
\pgftext[x=4.786188in,y=3.104105in,left,base]{\color{textcolor}\rmfamily\fontsize{10.000000}{12.000000}\selectfont No-Risk}%
\end{pgfscope}%
\begin{pgfscope}%
\pgfsetbuttcap%
\pgfsetroundjoin%
\pgfsetlinewidth{1.505625pt}%
\definecolor{currentstroke}{rgb}{0.678431,0.847059,0.901961}%
\pgfsetstrokecolor{currentstroke}%
\pgfsetdash{{5.550000pt}{2.400000pt}}{0.000000pt}%
\pgfpathmoveto{\pgfqpoint{4.397299in}{2.959043in}}%
\pgfpathlineto{\pgfqpoint{4.675076in}{2.959043in}}%
\pgfusepath{stroke}%
\end{pgfscope}%
\begin{pgfscope}%
\definecolor{textcolor}{rgb}{0.000000,0.000000,0.000000}%
\pgfsetstrokecolor{textcolor}%
\pgfsetfillcolor{textcolor}%
\pgftext[x=4.786188in,y=2.910432in,left,base]{\color{textcolor}\rmfamily\fontsize{10.000000}{12.000000}\selectfont All}%
\end{pgfscope}%
\end{pgfpicture}%
\makeatother%
\endgroup%

%% file: LaTeX/stats.tex
\subsection{Static Analysis}

We consider mean \textit{Compressibility} over the entire listening history of an individual and run a two-tailed Mann Whitney-U (MWU) test between the At-Risk and No-Risk groups. No significant differences were observed based on this test. We also consider the top \textit{n} $(100, 250, 500)$ songs of a user, based on playcount and observed no significant results.

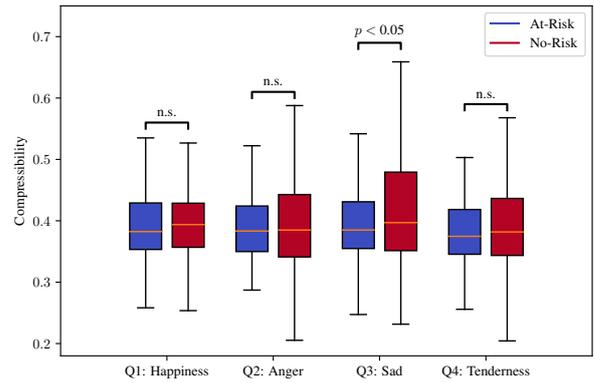
\begin{figure}[t]
  \centering  
  \resizebox{\linewidth}{!}{\input{LaTeX/images/quadrant_boxplot_compress.pgf}}
  \caption{Boxplots for mean \textit{Compressibility} per emotion quadrant \cite{russel1980}.\; \textit{n.s - not significant}} 
  \label{fig:quadrant_compress}
\end{figure}

\begin{figure}[t]
  \centering  
  \resizebox{\linewidth}{!}{\input{LaTeX/images/quadrant_boxplot_info.pgf}}
  \caption{Boxplots for mean \textit{AIC} per emotion quadrant \cite{russel1980}.\; \textit{n.s - not significant}} 
  \label{fig:quadrant_aic}
\end{figure}
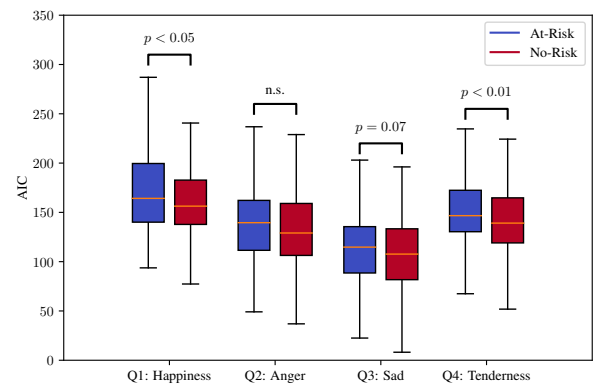

\begin{table}[h]
  \caption{MWU Test Results for mean AIC over the top $n$ tracks between the At-Risk and No-Risk groups}
  \label{tab:aic_table}
  \centering
  \begin{tabular}{ l  r  r }
    \toprule
    \multicolumn{1}{c}{\textbf{Number of Tracks}} & 
                            \multicolumn{1}{c}{\textbf{$U$ statistic}} & 
                            \multicolumn{1}{c}{\textbf{$p$ value}} \\
    \midrule
    $100$ & 12460 & 0.078\\
    $250$ & 12460 & 0.056\\
    $500$ & 12071 & 0.031\\
    All & 12117 & 0.035 \\
    \bottomrule
  \end{tabular}
  
\end{table}

However, when considering the AIC metric, the MWU test showed a significant difference for the top $n=500$ and entire listening history ($p<0.05$) and a borderline significant difference for the top $n=100, 250$, with a higher median for the At-Risk group in all cases.  Table \ref{tab:aic_table} summarises the MWU results and Figure \ref{fig:info_all} shows the median \textit{Compressiblity} per group for each number of tracks. 

Quadrant-specific analysis of \textit{Compressibility} revealed the At-Risk group have a significantly lower median in the quadrant representing \textit{Sadness} $(p < 0.05)$, with no significant difference for the other emotions. Figure \ref{fig:quadrant_compress} summarises this result. We also conduct a Spearman correlation between the valence and compressibility scores for all songs in the dataset, finding a weak but significant positive relationship ($r = 0.0756$, $p < 0.001$).

Additionally, quadrant-specific analysis of AIC revealed that songs listened to by the At-Risk group have a higher median in positively valenced quadrants \textit{Happiness} ($p<0.05$), \textit{Tenderness} ($p<0.01)$. There is also a borderline significant difference for \textit{Sadness} ($p=0.07$), with a higher median for the At-Risk group. Figure \ref{fig:quadrant_aic} summarises this result.

The ANOVA revealed no significant effect for \textit{compressibility}, while a significant main effect of age and interaction ($p<0.01$) was observed for AIC, with a higher AIC for the No Risk group in the 15-25 age range.

\subsection{Dynamic Analysis}

No significant differences were observed for intra and inter-session variability as per the MWU test for \textit{Compressibility}. Intra-session and Inter-session variability for AIC showed a significant difference ($p<0.01$, $p<0.05$ respectively) between the groups, with a higher median for both for the At-Risk group.

%% file: LaTeX/images/quadrant_boxplot_compress.pgf
\begingroup%
\makeatletter%
\begin{pgfpicture}%
\pgfpathrectangle{\pgfpointorigin}{\pgfqpoint{6.000000in}{4.000000in}}%
\pgfusepath{use as bounding box, clip}%
\begin{pgfscope}%
\pgfsetbuttcap%
\pgfsetmiterjoin%
\pgfsetlinewidth{0.000000pt}%
\definecolor{currentstroke}{rgb}{1.000000,1.000000,1.000000}%
\pgfsetstrokecolor{currentstroke}%
\pgfsetstrokeopacity{0.000000}%
\pgfsetdash{}{0pt}%
\pgfpathmoveto{\pgfqpoint{0.000000in}{0.000000in}}%
\pgfpathlineto{\pgfqpoint{6.000000in}{0.000000in}}%
\pgfpathlineto{\pgfqpoint{6.000000in}{4.000000in}}%
\pgfpathlineto{\pgfqpoint{0.000000in}{4.000000in}}%
\pgfpathclose%
\pgfusepath{}%
\end{pgfscope}%
\begin{pgfscope}%
\pgfsetbuttcap%
\pgfsetmiterjoin%
\definecolor{currentfill}{rgb}{1.000000,1.000000,1.000000}%
\pgfsetfillcolor{currentfill}%
\pgfsetlinewidth{0.000000pt}%
\definecolor{currentstroke}{rgb}{0.000000,0.000000,0.000000}%
\pgfsetstrokecolor{currentstroke}%
\pgfsetstrokeopacity{0.000000}%
\pgfsetdash{}{0pt}%
\pgfpathmoveto{\pgfqpoint{0.603704in}{0.370679in}}%
\pgfpathlineto{\pgfqpoint{5.850000in}{0.370679in}}%
\pgfpathlineto{\pgfqpoint{5.850000in}{3.850000in}}%
\pgfpathlineto{\pgfqpoint{0.603704in}{3.850000in}}%
\pgfpathclose%
\pgfusepath{fill}%
\end{pgfscope}%
\begin{pgfscope}%
\pgfsetbuttcap%
\pgfsetroundjoin%
\definecolor{currentfill}{rgb}{0.000000,0.000000,0.000000}%
\pgfsetfillcolor{currentfill}%
\pgfsetlinewidth{0.803000pt}%
\definecolor{currentstroke}{rgb}{0.000000,0.000000,0.000000}%
\pgfsetstrokecolor{currentstroke}%
\pgfsetdash{}{0pt}%
\pgfsys@defobject{currentmarker}{\pgfqpoint{0.000000in}{-0.048611in}}{\pgfqpoint{0.000000in}{0.000000in}}{%
\pgfpathmoveto{\pgfqpoint{0.000000in}{0.000000in}}%
\pgfpathlineto{\pgfqpoint{0.000000in}{-0.048611in}}%
\pgfusepath{stroke,fill}%
}%
\begin{pgfscope}%
\pgfsys@transformshift{1.652963in}{0.370679in}%
\pgfsys@useobject{currentmarker}{}%
\end{pgfscope}%
\end{pgfscope}%
\begin{pgfscope}%
\definecolor{textcolor}{rgb}{0.000000,0.000000,0.000000}%
\pgfsetstrokecolor{textcolor}%
\pgfsetfillcolor{textcolor}%
\pgftext[x=1.652963in,y=0.273457in,,top]{\color{textcolor}\rmfamily\fontsize{10.000000}{12.000000}\selectfont Q1: Happiness}%
\end{pgfscope}%
\begin{pgfscope}%
\pgfsetbuttcap%
\pgfsetroundjoin%
\definecolor{currentfill}{rgb}{0.000000,0.000000,0.000000}%
\pgfsetfillcolor{currentfill}%
\pgfsetlinewidth{0.803000pt}%
\definecolor{currentstroke}{rgb}{0.000000,0.000000,0.000000}%
\pgfsetstrokecolor{currentstroke}%
\pgfsetdash{}{0pt}%
\pgfsys@defobject{currentmarker}{\pgfqpoint{0.000000in}{-0.048611in}}{\pgfqpoint{0.000000in}{0.000000in}}{%
\pgfpathmoveto{\pgfqpoint{0.000000in}{0.000000in}}%
\pgfpathlineto{\pgfqpoint{0.000000in}{-0.048611in}}%
\pgfusepath{stroke,fill}%
}%
\begin{pgfscope}%
\pgfsys@transformshift{2.702222in}{0.370679in}%
\pgfsys@useobject{currentmarker}{}%
\end{pgfscope}%
\end{pgfscope}%
\begin{pgfscope}%
\definecolor{textcolor}{rgb}{0.000000,0.000000,0.000000}%
\pgfsetstrokecolor{textcolor}%
\pgfsetfillcolor{textcolor}%
\pgftext[x=2.702222in,y=0.273457in,,top]{\color{textcolor}\rmfamily\fontsize{10.000000}{12.000000}\selectfont Q2: Anger}%
\end{pgfscope}%
\begin{pgfscope}%
\pgfsetbuttcap%
\pgfsetroundjoin%
\definecolor{currentfill}{rgb}{0.000000,0.000000,0.000000}%
\pgfsetfillcolor{currentfill}%
\pgfsetlinewidth{0.803000pt}%
\definecolor{currentstroke}{rgb}{0.000000,0.000000,0.000000}%
\pgfsetstrokecolor{currentstroke}%
\pgfsetdash{}{0pt}%
\pgfsys@defobject{currentmarker}{\pgfqpoint{0.000000in}{-0.048611in}}{\pgfqpoint{0.000000in}{0.000000in}}{%
\pgfpathmoveto{\pgfqpoint{0.000000in}{0.000000in}}%
\pgfpathlineto{\pgfqpoint{0.000000in}{-0.048611in}}%
\pgfusepath{stroke,fill}%
}%
\begin{pgfscope}%
\pgfsys@transformshift{3.751482in}{0.370679in}%
\pgfsys@useobject{currentmarker}{}%
\end{pgfscope}%
\end{pgfscope}%
\begin{pgfscope}%
\definecolor{textcolor}{rgb}{0.000000,0.000000,0.000000}%
\pgfsetstrokecolor{textcolor}%
\pgfsetfillcolor{textcolor}%
\pgftext[x=3.751482in,y=0.273457in,,top]{\color{textcolor}\rmfamily\fontsize{10.000000}{12.000000}\selectfont Q3: Sad}%
\end{pgfscope}%
\begin{pgfscope}%
\pgfsetbuttcap%
\pgfsetroundjoin%
\definecolor{currentfill}{rgb}{0.000000,0.000000,0.000000}%
\pgfsetfillcolor{currentfill}%
\pgfsetlinewidth{0.803000pt}%
\definecolor{currentstroke}{rgb}{0.000000,0.000000,0.000000}%
\pgfsetstrokecolor{currentstroke}%
\pgfsetdash{}{0pt}%
\pgfsys@defobject{currentmarker}{\pgfqpoint{0.000000in}{-0.048611in}}{\pgfqpoint{0.000000in}{0.000000in}}{%
\pgfpathmoveto{\pgfqpoint{0.000000in}{0.000000in}}%
\pgfpathlineto{\pgfqpoint{0.000000in}{-0.048611in}}%
\pgfusepath{stroke,fill}%
}%
\begin{pgfscope}%
\pgfsys@transformshift{4.800741in}{0.370679in}%
\pgfsys@useobject{currentmarker}{}%
\end{pgfscope}%
\end{pgfscope}%
\begin{pgfscope}%
\definecolor{textcolor}{rgb}{0.000000,0.000000,0.000000}%
\pgfsetstrokecolor{textcolor}%
\pgfsetfillcolor{textcolor}%
\pgftext[x=4.800741in,y=0.273457in,,top]{\color{textcolor}\rmfamily\fontsize{10.000000}{12.000000}\selectfont Q4: Tenderness}%
\end{pgfscope}%
\begin{pgfscope}%
\pgfsetbuttcap%
\pgfsetroundjoin%
\definecolor{currentfill}{rgb}{0.000000,0.000000,0.000000}%
\pgfsetfillcolor{currentfill}%
\pgfsetlinewidth{0.803000pt}%
\definecolor{currentstroke}{rgb}{0.000000,0.000000,0.000000}%
\pgfsetstrokecolor{currentstroke}%
\pgfsetdash{}{0pt}%
\pgfsys@defobject{currentmarker}{\pgfqpoint{-0.048611in}{0.000000in}}{\pgfqpoint{-0.000000in}{0.000000in}}{%
\pgfpathmoveto{\pgfqpoint{-0.000000in}{0.000000in}}%
\pgfpathlineto{\pgfqpoint{-0.048611in}{0.000000in}}%
\pgfusepath{stroke,fill}%
}%
\begin{pgfscope}%
\pgfsys@transformshift{0.603704in}{0.492760in}%
\pgfsys@useobject{currentmarker}{}%
\end{pgfscope}%
\end{pgfscope}%
\begin{pgfscope}%
\definecolor{textcolor}{rgb}{0.000000,0.000000,0.000000}%
\pgfsetstrokecolor{textcolor}%
\pgfsetfillcolor{textcolor}%
\pgftext[x=0.329012in, y=0.444535in, left, base]{\color{textcolor}\rmfamily\fontsize{10.000000}{12.000000}\selectfont \(\displaystyle {0.2}\)}%
\end{pgfscope}%
\begin{pgfscope}%
\pgfsetbuttcap%
\pgfsetroundjoin%
\definecolor{currentfill}{rgb}{0.000000,0.000000,0.000000}%
\pgfsetfillcolor{currentfill}%
\pgfsetlinewidth{0.803000pt}%
\definecolor{currentstroke}{rgb}{0.000000,0.000000,0.000000}%
\pgfsetstrokecolor{currentstroke}%
\pgfsetdash{}{0pt}%
\pgfsys@defobject{currentmarker}{\pgfqpoint{-0.048611in}{0.000000in}}{\pgfqpoint{-0.000000in}{0.000000in}}{%
\pgfpathmoveto{\pgfqpoint{-0.000000in}{0.000000in}}%
\pgfpathlineto{\pgfqpoint{-0.048611in}{0.000000in}}%
\pgfusepath{stroke,fill}%
}%
\begin{pgfscope}%
\pgfsys@transformshift{0.603704in}{1.103168in}%
\pgfsys@useobject{currentmarker}{}%
\end{pgfscope}%
\end{pgfscope}%
\begin{pgfscope}%
\definecolor{textcolor}{rgb}{0.000000,0.000000,0.000000}%
\pgfsetstrokecolor{textcolor}%
\pgfsetfillcolor{textcolor}%
\pgftext[x=0.329012in, y=1.054942in, left, base]{\color{textcolor}\rmfamily\fontsize{10.000000}{12.000000}\selectfont \(\displaystyle {0.3}\)}%
\end{pgfscope}%
\begin{pgfscope}%
\pgfsetbuttcap%
\pgfsetroundjoin%
\definecolor{currentfill}{rgb}{0.000000,0.000000,0.000000}%
\pgfsetfillcolor{currentfill}%
\pgfsetlinewidth{0.803000pt}%
\definecolor{currentstroke}{rgb}{0.000000,0.000000,0.000000}%
\pgfsetstrokecolor{currentstroke}%
\pgfsetdash{}{0pt}%
\pgfsys@defobject{currentmarker}{\pgfqpoint{-0.048611in}{0.000000in}}{\pgfqpoint{-0.000000in}{0.000000in}}{%
\pgfpathmoveto{\pgfqpoint{-0.000000in}{0.000000in}}%
\pgfpathlineto{\pgfqpoint{-0.048611in}{0.000000in}}%
\pgfusepath{stroke,fill}%
}%
\begin{pgfscope}%
\pgfsys@transformshift{0.603704in}{1.713575in}%
\pgfsys@useobject{currentmarker}{}%
\end{pgfscope}%
\end{pgfscope}%
\begin{pgfscope}%
\definecolor{textcolor}{rgb}{0.000000,0.000000,0.000000}%
\pgfsetstrokecolor{textcolor}%
\pgfsetfillcolor{textcolor}%
\pgftext[x=0.329012in, y=1.665349in, left, base]{\color{textcolor}\rmfamily\fontsize{10.000000}{12.000000}\selectfont \(\displaystyle {0.4}\)}%
\end{pgfscope}%
\begin{pgfscope}%
\pgfsetbuttcap%
\pgfsetroundjoin%
\definecolor{currentfill}{rgb}{0.000000,0.000000,0.000000}%
\pgfsetfillcolor{currentfill}%
\pgfsetlinewidth{0.803000pt}%
\definecolor{currentstroke}{rgb}{0.000000,0.000000,0.000000}%
\pgfsetstrokecolor{currentstroke}%
\pgfsetdash{}{0pt}%
\pgfsys@defobject{currentmarker}{\pgfqpoint{-0.048611in}{0.000000in}}{\pgfqpoint{-0.000000in}{0.000000in}}{%
\pgfpathmoveto{\pgfqpoint{-0.000000in}{0.000000in}}%
\pgfpathlineto{\pgfqpoint{-0.048611in}{0.000000in}}%
\pgfusepath{stroke,fill}%
}%
\begin{pgfscope}%
\pgfsys@transformshift{0.603704in}{2.323982in}%
\pgfsys@useobject{currentmarker}{}%
\end{pgfscope}%
\end{pgfscope}%
\begin{pgfscope}%
\definecolor{textcolor}{rgb}{0.000000,0.000000,0.000000}%
\pgfsetstrokecolor{textcolor}%
\pgfsetfillcolor{textcolor}%
\pgftext[x=0.329012in, y=2.275757in, left, base]{\color{textcolor}\rmfamily\fontsize{10.000000}{12.000000}\selectfont \(\displaystyle {0.5}\)}%
\end{pgfscope}%
\begin{pgfscope}%
\pgfsetbuttcap%
\pgfsetroundjoin%
\definecolor{currentfill}{rgb}{0.000000,0.000000,0.000000}%
\pgfsetfillcolor{currentfill}%
\pgfsetlinewidth{0.803000pt}%
\definecolor{currentstroke}{rgb}{0.000000,0.000000,0.000000}%
\pgfsetstrokecolor{currentstroke}%
\pgfsetdash{}{0pt}%
\pgfsys@defobject{currentmarker}{\pgfqpoint{-0.048611in}{0.000000in}}{\pgfqpoint{-0.000000in}{0.000000in}}{%
\pgfpathmoveto{\pgfqpoint{-0.000000in}{0.000000in}}%
\pgfpathlineto{\pgfqpoint{-0.048611in}{0.000000in}}%
\pgfusepath{stroke,fill}%
}%
\begin{pgfscope}%
\pgfsys@transformshift{0.603704in}{2.934389in}%
\pgfsys@useobject{currentmarker}{}%
\end{pgfscope}%
\end{pgfscope}%
\begin{pgfscope}%
\definecolor{textcolor}{rgb}{0.000000,0.000000,0.000000}%
\pgfsetstrokecolor{textcolor}%
\pgfsetfillcolor{textcolor}%
\pgftext[x=0.329012in, y=2.886164in, left, base]{\color{textcolor}\rmfamily\fontsize{10.000000}{12.000000}\selectfont \(\displaystyle {0.6}\)}%
\end{pgfscope}%
\begin{pgfscope}%
\pgfsetbuttcap%
\pgfsetroundjoin%
\definecolor{currentfill}{rgb}{0.000000,0.000000,0.000000}%
\pgfsetfillcolor{currentfill}%
\pgfsetlinewidth{0.803000pt}%
\definecolor{currentstroke}{rgb}{0.000000,0.000000,0.000000}%
\pgfsetstrokecolor{currentstroke}%
\pgfsetdash{}{0pt}%
\pgfsys@defobject{currentmarker}{\pgfqpoint{-0.048611in}{0.000000in}}{\pgfqpoint{-0.000000in}{0.000000in}}{%
\pgfpathmoveto{\pgfqpoint{-0.000000in}{0.000000in}}%
\pgfpathlineto{\pgfqpoint{-0.048611in}{0.000000in}}%
\pgfusepath{stroke,fill}%
}%
\begin{pgfscope}%
\pgfsys@transformshift{0.603704in}{3.544796in}%
\pgfsys@useobject{currentmarker}{}%
\end{pgfscope}%
\end{pgfscope}%
\begin{pgfscope}%
\definecolor{textcolor}{rgb}{0.000000,0.000000,0.000000}%
\pgfsetstrokecolor{textcolor}%
\pgfsetfillcolor{textcolor}%
\pgftext[x=0.329012in, y=3.496571in, left, base]{\color{textcolor}\rmfamily\fontsize{10.000000}{12.000000}\selectfont \(\displaystyle {0.7}\)}%
\end{pgfscope}%
\begin{pgfscope}%
\definecolor{textcolor}{rgb}{0.000000,0.000000,0.000000}%
\pgfsetstrokecolor{textcolor}%
\pgfsetfillcolor{textcolor}%
\pgftext[x=0.273457in,y=2.110339in,,bottom,rotate=90.000000]{\color{textcolor}\rmfamily\fontsize{10.000000}{12.000000}\selectfont Compressibility}%
\end{pgfscope}%
\begin{pgfscope}%
\pgfpathrectangle{\pgfqpoint{0.603704in}{0.370679in}}{\pgfqpoint{5.246296in}{3.479321in}}%
\pgfusepath{clip}%
\pgfsetrectcap%
\pgfsetroundjoin%
\pgfsetlinewidth{1.003750pt}%
\definecolor{currentstroke}{rgb}{0.000000,0.000000,0.000000}%
\pgfsetstrokecolor{currentstroke}%
\pgfsetdash{}{0pt}%
\pgfpathmoveto{\pgfqpoint{1.443111in}{1.428612in}}%
\pgfpathlineto{\pgfqpoint{1.443111in}{0.848301in}}%
\pgfusepath{stroke}%
\end{pgfscope}%
\begin{pgfscope}%
\pgfpathrectangle{\pgfqpoint{0.603704in}{0.370679in}}{\pgfqpoint{5.246296in}{3.479321in}}%
\pgfusepath{clip}%
\pgfsetrectcap%
\pgfsetroundjoin%
\pgfsetlinewidth{1.003750pt}%
\definecolor{currentstroke}{rgb}{0.000000,0.000000,0.000000}%
\pgfsetstrokecolor{currentstroke}%
\pgfsetdash{}{0pt}%
\pgfpathmoveto{\pgfqpoint{1.443111in}{1.890833in}}%
\pgfpathlineto{\pgfqpoint{1.443111in}{2.537657in}}%
\pgfusepath{stroke}%
\end{pgfscope}%
\begin{pgfscope}%
\pgfpathrectangle{\pgfqpoint{0.603704in}{0.370679in}}{\pgfqpoint{5.246296in}{3.479321in}}%
\pgfusepath{clip}%
\pgfsetrectcap%
\pgfsetroundjoin%
\pgfsetlinewidth{1.003750pt}%
\definecolor{currentstroke}{rgb}{0.000000,0.000000,0.000000}%
\pgfsetstrokecolor{currentstroke}%
\pgfsetdash{}{0pt}%
\pgfpathmoveto{\pgfqpoint{1.364417in}{0.848301in}}%
\pgfpathlineto{\pgfqpoint{1.521806in}{0.848301in}}%
\pgfusepath{stroke}%
\end{pgfscope}%
\begin{pgfscope}%
\pgfpathrectangle{\pgfqpoint{0.603704in}{0.370679in}}{\pgfqpoint{5.246296in}{3.479321in}}%
\pgfusepath{clip}%
\pgfsetrectcap%
\pgfsetroundjoin%
\pgfsetlinewidth{1.003750pt}%
\definecolor{currentstroke}{rgb}{0.000000,0.000000,0.000000}%
\pgfsetstrokecolor{currentstroke}%
\pgfsetdash{}{0pt}%
\pgfpathmoveto{\pgfqpoint{1.364417in}{2.537657in}}%
\pgfpathlineto{\pgfqpoint{1.521806in}{2.537657in}}%
\pgfusepath{stroke}%
\end{pgfscope}%
\begin{pgfscope}%
\pgfpathrectangle{\pgfqpoint{0.603704in}{0.370679in}}{\pgfqpoint{5.246296in}{3.479321in}}%
\pgfusepath{clip}%
\pgfsetrectcap%
\pgfsetroundjoin%
\pgfsetlinewidth{1.003750pt}%
\definecolor{currentstroke}{rgb}{0.000000,0.000000,0.000000}%
\pgfsetstrokecolor{currentstroke}%
\pgfsetdash{}{0pt}%
\pgfpathmoveto{\pgfqpoint{2.492371in}{1.407582in}}%
\pgfpathlineto{\pgfqpoint{2.492371in}{1.024551in}}%
\pgfusepath{stroke}%
\end{pgfscope}%
\begin{pgfscope}%
\pgfpathrectangle{\pgfqpoint{0.603704in}{0.370679in}}{\pgfqpoint{5.246296in}{3.479321in}}%
\pgfusepath{clip}%
\pgfsetrectcap%
\pgfsetroundjoin%
\pgfsetlinewidth{1.003750pt}%
\definecolor{currentstroke}{rgb}{0.000000,0.000000,0.000000}%
\pgfsetstrokecolor{currentstroke}%
\pgfsetdash{}{0pt}%
\pgfpathmoveto{\pgfqpoint{2.492371in}{1.860426in}}%
\pgfpathlineto{\pgfqpoint{2.492371in}{2.459763in}}%
\pgfusepath{stroke}%
\end{pgfscope}%
\begin{pgfscope}%
\pgfpathrectangle{\pgfqpoint{0.603704in}{0.370679in}}{\pgfqpoint{5.246296in}{3.479321in}}%
\pgfusepath{clip}%
\pgfsetrectcap%
\pgfsetroundjoin%
\pgfsetlinewidth{1.003750pt}%
\definecolor{currentstroke}{rgb}{0.000000,0.000000,0.000000}%
\pgfsetstrokecolor{currentstroke}%
\pgfsetdash{}{0pt}%
\pgfpathmoveto{\pgfqpoint{2.413676in}{1.024551in}}%
\pgfpathlineto{\pgfqpoint{2.571065in}{1.024551in}}%
\pgfusepath{stroke}%
\end{pgfscope}%
\begin{pgfscope}%
\pgfpathrectangle{\pgfqpoint{0.603704in}{0.370679in}}{\pgfqpoint{5.246296in}{3.479321in}}%
\pgfusepath{clip}%
\pgfsetrectcap%
\pgfsetroundjoin%
\pgfsetlinewidth{1.003750pt}%
\definecolor{currentstroke}{rgb}{0.000000,0.000000,0.000000}%
\pgfsetstrokecolor{currentstroke}%
\pgfsetdash{}{0pt}%
\pgfpathmoveto{\pgfqpoint{2.413676in}{2.459763in}}%
\pgfpathlineto{\pgfqpoint{2.571065in}{2.459763in}}%
\pgfusepath{stroke}%
\end{pgfscope}%
\begin{pgfscope}%
\pgfpathrectangle{\pgfqpoint{0.603704in}{0.370679in}}{\pgfqpoint{5.246296in}{3.479321in}}%
\pgfusepath{clip}%
\pgfsetrectcap%
\pgfsetroundjoin%
\pgfsetlinewidth{1.003750pt}%
\definecolor{currentstroke}{rgb}{0.000000,0.000000,0.000000}%
\pgfsetstrokecolor{currentstroke}%
\pgfsetdash{}{0pt}%
\pgfpathmoveto{\pgfqpoint{3.541630in}{1.437930in}}%
\pgfpathlineto{\pgfqpoint{3.541630in}{0.781630in}}%
\pgfusepath{stroke}%
\end{pgfscope}%
\begin{pgfscope}%
\pgfpathrectangle{\pgfqpoint{0.603704in}{0.370679in}}{\pgfqpoint{5.246296in}{3.479321in}}%
\pgfusepath{clip}%
\pgfsetrectcap%
\pgfsetroundjoin%
\pgfsetlinewidth{1.003750pt}%
\definecolor{currentstroke}{rgb}{0.000000,0.000000,0.000000}%
\pgfsetstrokecolor{currentstroke}%
\pgfsetdash{}{0pt}%
\pgfpathmoveto{\pgfqpoint{3.541630in}{1.902611in}}%
\pgfpathlineto{\pgfqpoint{3.541630in}{2.578812in}}%
\pgfusepath{stroke}%
\end{pgfscope}%
\begin{pgfscope}%
\pgfpathrectangle{\pgfqpoint{0.603704in}{0.370679in}}{\pgfqpoint{5.246296in}{3.479321in}}%
\pgfusepath{clip}%
\pgfsetrectcap%
\pgfsetroundjoin%
\pgfsetlinewidth{1.003750pt}%
\definecolor{currentstroke}{rgb}{0.000000,0.000000,0.000000}%
\pgfsetstrokecolor{currentstroke}%
\pgfsetdash{}{0pt}%
\pgfpathmoveto{\pgfqpoint{3.462935in}{0.781630in}}%
\pgfpathlineto{\pgfqpoint{3.620324in}{0.781630in}}%
\pgfusepath{stroke}%
\end{pgfscope}%
\begin{pgfscope}%
\pgfpathrectangle{\pgfqpoint{0.603704in}{0.370679in}}{\pgfqpoint{5.246296in}{3.479321in}}%
\pgfusepath{clip}%
\pgfsetrectcap%
\pgfsetroundjoin%
\pgfsetlinewidth{1.003750pt}%
\definecolor{currentstroke}{rgb}{0.000000,0.000000,0.000000}%
\pgfsetstrokecolor{currentstroke}%
\pgfsetdash{}{0pt}%
\pgfpathmoveto{\pgfqpoint{3.462935in}{2.578812in}}%
\pgfpathlineto{\pgfqpoint{3.620324in}{2.578812in}}%
\pgfusepath{stroke}%
\end{pgfscope}%
\begin{pgfscope}%
\pgfpathrectangle{\pgfqpoint{0.603704in}{0.370679in}}{\pgfqpoint{5.246296in}{3.479321in}}%
\pgfusepath{clip}%
\pgfsetrectcap%
\pgfsetroundjoin%
\pgfsetlinewidth{1.003750pt}%
\definecolor{currentstroke}{rgb}{0.000000,0.000000,0.000000}%
\pgfsetstrokecolor{currentstroke}%
\pgfsetdash{}{0pt}%
\pgfpathmoveto{\pgfqpoint{4.590889in}{1.380629in}}%
\pgfpathlineto{\pgfqpoint{4.590889in}{0.833502in}}%
\pgfusepath{stroke}%
\end{pgfscope}%
\begin{pgfscope}%
\pgfpathrectangle{\pgfqpoint{0.603704in}{0.370679in}}{\pgfqpoint{5.246296in}{3.479321in}}%
\pgfusepath{clip}%
\pgfsetrectcap%
\pgfsetroundjoin%
\pgfsetlinewidth{1.003750pt}%
\definecolor{currentstroke}{rgb}{0.000000,0.000000,0.000000}%
\pgfsetstrokecolor{currentstroke}%
\pgfsetdash{}{0pt}%
\pgfpathmoveto{\pgfqpoint{4.590889in}{1.825902in}}%
\pgfpathlineto{\pgfqpoint{4.590889in}{2.342513in}}%
\pgfusepath{stroke}%
\end{pgfscope}%
\begin{pgfscope}%
\pgfpathrectangle{\pgfqpoint{0.603704in}{0.370679in}}{\pgfqpoint{5.246296in}{3.479321in}}%
\pgfusepath{clip}%
\pgfsetrectcap%
\pgfsetroundjoin%
\pgfsetlinewidth{1.003750pt}%
\definecolor{currentstroke}{rgb}{0.000000,0.000000,0.000000}%
\pgfsetstrokecolor{currentstroke}%
\pgfsetdash{}{0pt}%
\pgfpathmoveto{\pgfqpoint{4.512195in}{0.833502in}}%
\pgfpathlineto{\pgfqpoint{4.669583in}{0.833502in}}%
\pgfusepath{stroke}%
\end{pgfscope}%
\begin{pgfscope}%
\pgfpathrectangle{\pgfqpoint{0.603704in}{0.370679in}}{\pgfqpoint{5.246296in}{3.479321in}}%
\pgfusepath{clip}%
\pgfsetrectcap%
\pgfsetroundjoin%
\pgfsetlinewidth{1.003750pt}%
\definecolor{currentstroke}{rgb}{0.000000,0.000000,0.000000}%
\pgfsetstrokecolor{currentstroke}%
\pgfsetdash{}{0pt}%
\pgfpathmoveto{\pgfqpoint{4.512195in}{2.342513in}}%
\pgfpathlineto{\pgfqpoint{4.669583in}{2.342513in}}%
\pgfusepath{stroke}%
\end{pgfscope}%
\begin{pgfscope}%
\pgfpathrectangle{\pgfqpoint{0.603704in}{0.370679in}}{\pgfqpoint{5.246296in}{3.479321in}}%
\pgfusepath{clip}%
\pgfsetrectcap%
\pgfsetroundjoin%
\pgfsetlinewidth{1.003750pt}%
\definecolor{currentstroke}{rgb}{0.000000,0.000000,0.000000}%
\pgfsetstrokecolor{currentstroke}%
\pgfsetdash{}{0pt}%
\pgfpathmoveto{\pgfqpoint{1.862815in}{1.450011in}}%
\pgfpathlineto{\pgfqpoint{1.862815in}{0.820219in}}%
\pgfusepath{stroke}%
\end{pgfscope}%
\begin{pgfscope}%
\pgfpathrectangle{\pgfqpoint{0.603704in}{0.370679in}}{\pgfqpoint{5.246296in}{3.479321in}}%
\pgfusepath{clip}%
\pgfsetrectcap%
\pgfsetroundjoin%
\pgfsetlinewidth{1.003750pt}%
\definecolor{currentstroke}{rgb}{0.000000,0.000000,0.000000}%
\pgfsetstrokecolor{currentstroke}%
\pgfsetdash{}{0pt}%
\pgfpathmoveto{\pgfqpoint{1.862815in}{1.888473in}}%
\pgfpathlineto{\pgfqpoint{1.862815in}{2.486906in}}%
\pgfusepath{stroke}%
\end{pgfscope}%
\begin{pgfscope}%
\pgfpathrectangle{\pgfqpoint{0.603704in}{0.370679in}}{\pgfqpoint{5.246296in}{3.479321in}}%
\pgfusepath{clip}%
\pgfsetrectcap%
\pgfsetroundjoin%
\pgfsetlinewidth{1.003750pt}%
\definecolor{currentstroke}{rgb}{0.000000,0.000000,0.000000}%
\pgfsetstrokecolor{currentstroke}%
\pgfsetdash{}{0pt}%
\pgfpathmoveto{\pgfqpoint{1.784121in}{0.820219in}}%
\pgfpathlineto{\pgfqpoint{1.941510in}{0.820219in}}%
\pgfusepath{stroke}%
\end{pgfscope}%
\begin{pgfscope}%
\pgfpathrectangle{\pgfqpoint{0.603704in}{0.370679in}}{\pgfqpoint{5.246296in}{3.479321in}}%
\pgfusepath{clip}%
\pgfsetrectcap%
\pgfsetroundjoin%
\pgfsetlinewidth{1.003750pt}%
\definecolor{currentstroke}{rgb}{0.000000,0.000000,0.000000}%
\pgfsetstrokecolor{currentstroke}%
\pgfsetdash{}{0pt}%
\pgfpathmoveto{\pgfqpoint{1.784121in}{2.486906in}}%
\pgfpathlineto{\pgfqpoint{1.941510in}{2.486906in}}%
\pgfusepath{stroke}%
\end{pgfscope}%
\begin{pgfscope}%
\pgfpathrectangle{\pgfqpoint{0.603704in}{0.370679in}}{\pgfqpoint{5.246296in}{3.479321in}}%
\pgfusepath{clip}%
\pgfsetrectcap%
\pgfsetroundjoin%
\pgfsetlinewidth{1.003750pt}%
\definecolor{currentstroke}{rgb}{0.000000,0.000000,0.000000}%
\pgfsetstrokecolor{currentstroke}%
\pgfsetdash{}{0pt}%
\pgfpathmoveto{\pgfqpoint{2.912074in}{1.353976in}}%
\pgfpathlineto{\pgfqpoint{2.912074in}{0.525200in}}%
\pgfusepath{stroke}%
\end{pgfscope}%
\begin{pgfscope}%
\pgfpathrectangle{\pgfqpoint{0.603704in}{0.370679in}}{\pgfqpoint{5.246296in}{3.479321in}}%
\pgfusepath{clip}%
\pgfsetrectcap%
\pgfsetroundjoin%
\pgfsetlinewidth{1.003750pt}%
\definecolor{currentstroke}{rgb}{0.000000,0.000000,0.000000}%
\pgfsetstrokecolor{currentstroke}%
\pgfsetdash{}{0pt}%
\pgfpathmoveto{\pgfqpoint{2.912074in}{1.974415in}}%
\pgfpathlineto{\pgfqpoint{2.912074in}{2.859171in}}%
\pgfusepath{stroke}%
\end{pgfscope}%
\begin{pgfscope}%
\pgfpathrectangle{\pgfqpoint{0.603704in}{0.370679in}}{\pgfqpoint{5.246296in}{3.479321in}}%
\pgfusepath{clip}%
\pgfsetrectcap%
\pgfsetroundjoin%
\pgfsetlinewidth{1.003750pt}%
\definecolor{currentstroke}{rgb}{0.000000,0.000000,0.000000}%
\pgfsetstrokecolor{currentstroke}%
\pgfsetdash{}{0pt}%
\pgfpathmoveto{\pgfqpoint{2.833380in}{0.525200in}}%
\pgfpathlineto{\pgfqpoint{2.990769in}{0.525200in}}%
\pgfusepath{stroke}%
\end{pgfscope}%
\begin{pgfscope}%
\pgfpathrectangle{\pgfqpoint{0.603704in}{0.370679in}}{\pgfqpoint{5.246296in}{3.479321in}}%
\pgfusepath{clip}%
\pgfsetrectcap%
\pgfsetroundjoin%
\pgfsetlinewidth{1.003750pt}%
\definecolor{currentstroke}{rgb}{0.000000,0.000000,0.000000}%
\pgfsetstrokecolor{currentstroke}%
\pgfsetdash{}{0pt}%
\pgfpathmoveto{\pgfqpoint{2.833380in}{2.859171in}}%
\pgfpathlineto{\pgfqpoint{2.990769in}{2.859171in}}%
\pgfusepath{stroke}%
\end{pgfscope}%
\begin{pgfscope}%
\pgfpathrectangle{\pgfqpoint{0.603704in}{0.370679in}}{\pgfqpoint{5.246296in}{3.479321in}}%
\pgfusepath{clip}%
\pgfsetrectcap%
\pgfsetroundjoin%
\pgfsetlinewidth{1.003750pt}%
\definecolor{currentstroke}{rgb}{0.000000,0.000000,0.000000}%
\pgfsetstrokecolor{currentstroke}%
\pgfsetdash{}{0pt}%
\pgfpathmoveto{\pgfqpoint{3.961333in}{1.416760in}}%
\pgfpathlineto{\pgfqpoint{3.961333in}{0.685541in}}%
\pgfusepath{stroke}%
\end{pgfscope}%
\begin{pgfscope}%
\pgfpathrectangle{\pgfqpoint{0.603704in}{0.370679in}}{\pgfqpoint{5.246296in}{3.479321in}}%
\pgfusepath{clip}%
\pgfsetrectcap%
\pgfsetroundjoin%
\pgfsetlinewidth{1.003750pt}%
\definecolor{currentstroke}{rgb}{0.000000,0.000000,0.000000}%
\pgfsetstrokecolor{currentstroke}%
\pgfsetdash{}{0pt}%
\pgfpathmoveto{\pgfqpoint{3.961333in}{2.197422in}}%
\pgfpathlineto{\pgfqpoint{3.961333in}{3.294322in}}%
\pgfusepath{stroke}%
\end{pgfscope}%
\begin{pgfscope}%
\pgfpathrectangle{\pgfqpoint{0.603704in}{0.370679in}}{\pgfqpoint{5.246296in}{3.479321in}}%
\pgfusepath{clip}%
\pgfsetrectcap%
\pgfsetroundjoin%
\pgfsetlinewidth{1.003750pt}%
\definecolor{currentstroke}{rgb}{0.000000,0.000000,0.000000}%
\pgfsetstrokecolor{currentstroke}%
\pgfsetdash{}{0pt}%
\pgfpathmoveto{\pgfqpoint{3.882639in}{0.685541in}}%
\pgfpathlineto{\pgfqpoint{4.040028in}{0.685541in}}%
\pgfusepath{stroke}%
\end{pgfscope}%
\begin{pgfscope}%
\pgfpathrectangle{\pgfqpoint{0.603704in}{0.370679in}}{\pgfqpoint{5.246296in}{3.479321in}}%
\pgfusepath{clip}%
\pgfsetrectcap%
\pgfsetroundjoin%
\pgfsetlinewidth{1.003750pt}%
\definecolor{currentstroke}{rgb}{0.000000,0.000000,0.000000}%
\pgfsetstrokecolor{currentstroke}%
\pgfsetdash{}{0pt}%
\pgfpathmoveto{\pgfqpoint{3.882639in}{3.294322in}}%
\pgfpathlineto{\pgfqpoint{4.040028in}{3.294322in}}%
\pgfusepath{stroke}%
\end{pgfscope}%
\begin{pgfscope}%
\pgfpathrectangle{\pgfqpoint{0.603704in}{0.370679in}}{\pgfqpoint{5.246296in}{3.479321in}}%
\pgfusepath{clip}%
\pgfsetrectcap%
\pgfsetroundjoin%
\pgfsetlinewidth{1.003750pt}%
\definecolor{currentstroke}{rgb}{0.000000,0.000000,0.000000}%
\pgfsetstrokecolor{currentstroke}%
\pgfsetdash{}{0pt}%
\pgfpathmoveto{\pgfqpoint{5.010593in}{1.369016in}}%
\pgfpathlineto{\pgfqpoint{5.010593in}{0.519225in}}%
\pgfusepath{stroke}%
\end{pgfscope}%
\begin{pgfscope}%
\pgfpathrectangle{\pgfqpoint{0.603704in}{0.370679in}}{\pgfqpoint{5.246296in}{3.479321in}}%
\pgfusepath{clip}%
\pgfsetrectcap%
\pgfsetroundjoin%
\pgfsetlinewidth{1.003750pt}%
\definecolor{currentstroke}{rgb}{0.000000,0.000000,0.000000}%
\pgfsetstrokecolor{currentstroke}%
\pgfsetdash{}{0pt}%
\pgfpathmoveto{\pgfqpoint{5.010593in}{1.935875in}}%
\pgfpathlineto{\pgfqpoint{5.010593in}{2.738360in}}%
\pgfusepath{stroke}%
\end{pgfscope}%
\begin{pgfscope}%
\pgfpathrectangle{\pgfqpoint{0.603704in}{0.370679in}}{\pgfqpoint{5.246296in}{3.479321in}}%
\pgfusepath{clip}%
\pgfsetrectcap%
\pgfsetroundjoin%
\pgfsetlinewidth{1.003750pt}%
\definecolor{currentstroke}{rgb}{0.000000,0.000000,0.000000}%
\pgfsetstrokecolor{currentstroke}%
\pgfsetdash{}{0pt}%
\pgfpathmoveto{\pgfqpoint{4.931898in}{0.519225in}}%
\pgfpathlineto{\pgfqpoint{5.089287in}{0.519225in}}%
\pgfusepath{stroke}%
\end{pgfscope}%
\begin{pgfscope}%
\pgfpathrectangle{\pgfqpoint{0.603704in}{0.370679in}}{\pgfqpoint{5.246296in}{3.479321in}}%
\pgfusepath{clip}%
\pgfsetrectcap%
\pgfsetroundjoin%
\pgfsetlinewidth{1.003750pt}%
\definecolor{currentstroke}{rgb}{0.000000,0.000000,0.000000}%
\pgfsetstrokecolor{currentstroke}%
\pgfsetdash{}{0pt}%
\pgfpathmoveto{\pgfqpoint{4.931898in}{2.738360in}}%
\pgfpathlineto{\pgfqpoint{5.089287in}{2.738360in}}%
\pgfusepath{stroke}%
\end{pgfscope}%
\begin{pgfscope}%
\pgfpathrectangle{\pgfqpoint{0.603704in}{0.370679in}}{\pgfqpoint{5.246296in}{3.479321in}}%
\pgfusepath{clip}%
\pgfsetrectcap%
\pgfsetroundjoin%
\pgfsetlinewidth{1.505625pt}%
\definecolor{currentstroke}{rgb}{0.000000,0.000000,0.000000}%
\pgfsetstrokecolor{currentstroke}%
\pgfsetdash{}{0pt}%
\pgfpathmoveto{\pgfqpoint{1.443111in}{2.629186in}}%
\pgfpathlineto{\pgfqpoint{1.443111in}{2.690226in}}%
\pgfpathlineto{\pgfqpoint{1.862815in}{2.690226in}}%
\pgfpathlineto{\pgfqpoint{1.862815in}{2.629186in}}%
\pgfusepath{stroke}%
\end{pgfscope}%
\begin{pgfscope}%
\pgfpathrectangle{\pgfqpoint{0.603704in}{0.370679in}}{\pgfqpoint{5.246296in}{3.479321in}}%
\pgfusepath{clip}%
\pgfsetrectcap%
\pgfsetroundjoin%
\pgfsetlinewidth{1.505625pt}%
\definecolor{currentstroke}{rgb}{0.000000,0.000000,0.000000}%
\pgfsetstrokecolor{currentstroke}%
\pgfsetdash{}{0pt}%
\pgfpathmoveto{\pgfqpoint{2.492371in}{2.934389in}}%
\pgfpathlineto{\pgfqpoint{2.492371in}{2.995430in}}%
\pgfpathlineto{\pgfqpoint{2.912074in}{2.995430in}}%
\pgfpathlineto{\pgfqpoint{2.912074in}{2.934389in}}%
\pgfusepath{stroke}%
\end{pgfscope}%
\begin{pgfscope}%
\pgfpathrectangle{\pgfqpoint{0.603704in}{0.370679in}}{\pgfqpoint{5.246296in}{3.479321in}}%
\pgfusepath{clip}%
\pgfsetrectcap%
\pgfsetroundjoin%
\pgfsetlinewidth{1.505625pt}%
\definecolor{currentstroke}{rgb}{0.000000,0.000000,0.000000}%
\pgfsetstrokecolor{currentstroke}%
\pgfsetdash{}{0pt}%
\pgfpathmoveto{\pgfqpoint{3.541630in}{3.422715in}}%
\pgfpathlineto{\pgfqpoint{3.541630in}{3.483756in}}%
\pgfpathlineto{\pgfqpoint{3.961333in}{3.483756in}}%
\pgfpathlineto{\pgfqpoint{3.961333in}{3.422715in}}%
\pgfusepath{stroke}%
\end{pgfscope}%
\begin{pgfscope}%
\pgfpathrectangle{\pgfqpoint{0.603704in}{0.370679in}}{\pgfqpoint{5.246296in}{3.479321in}}%
\pgfusepath{clip}%
\pgfsetrectcap%
\pgfsetroundjoin%
\pgfsetlinewidth{1.505625pt}%
\definecolor{currentstroke}{rgb}{0.000000,0.000000,0.000000}%
\pgfsetstrokecolor{currentstroke}%
\pgfsetdash{}{0pt}%
\pgfpathmoveto{\pgfqpoint{4.590889in}{2.812308in}}%
\pgfpathlineto{\pgfqpoint{4.590889in}{2.873348in}}%
\pgfpathlineto{\pgfqpoint{5.010593in}{2.873348in}}%
\pgfpathlineto{\pgfqpoint{5.010593in}{2.812308in}}%
\pgfusepath{stroke}%
\end{pgfscope}%
\begin{pgfscope}%
\pgfpathrectangle{\pgfqpoint{0.603704in}{0.370679in}}{\pgfqpoint{5.246296in}{3.479321in}}%
\pgfusepath{clip}%
\pgfsetbuttcap%
\pgfsetmiterjoin%
\definecolor{currentfill}{rgb}{0.229806,0.298718,0.753683}%
\pgfsetfillcolor{currentfill}%
\pgfsetlinewidth{1.003750pt}%
\definecolor{currentstroke}{rgb}{0.000000,0.000000,0.000000}%
\pgfsetstrokecolor{currentstroke}%
\pgfsetdash{}{0pt}%
\pgfpathmoveto{\pgfqpoint{1.285723in}{1.428612in}}%
\pgfpathlineto{\pgfqpoint{1.600500in}{1.428612in}}%
\pgfpathlineto{\pgfqpoint{1.600500in}{1.890833in}}%
\pgfpathlineto{\pgfqpoint{1.285723in}{1.890833in}}%
\pgfpathlineto{\pgfqpoint{1.285723in}{1.428612in}}%
\pgfpathclose%
\pgfusepath{stroke,fill}%
\end{pgfscope}%
\begin{pgfscope}%
\pgfpathrectangle{\pgfqpoint{0.603704in}{0.370679in}}{\pgfqpoint{5.246296in}{3.479321in}}%
\pgfusepath{clip}%
\pgfsetbuttcap%
\pgfsetmiterjoin%
\definecolor{currentfill}{rgb}{0.229806,0.298718,0.753683}%
\pgfsetfillcolor{currentfill}%
\pgfsetlinewidth{1.003750pt}%
\definecolor{currentstroke}{rgb}{0.000000,0.000000,0.000000}%
\pgfsetstrokecolor{currentstroke}%
\pgfsetdash{}{0pt}%
\pgfpathmoveto{\pgfqpoint{2.334982in}{1.407582in}}%
\pgfpathlineto{\pgfqpoint{2.649760in}{1.407582in}}%
\pgfpathlineto{\pgfqpoint{2.649760in}{1.860426in}}%
\pgfpathlineto{\pgfqpoint{2.334982in}{1.860426in}}%
\pgfpathlineto{\pgfqpoint{2.334982in}{1.407582in}}%
\pgfpathclose%
\pgfusepath{stroke,fill}%
\end{pgfscope}%
\begin{pgfscope}%
\pgfpathrectangle{\pgfqpoint{0.603704in}{0.370679in}}{\pgfqpoint{5.246296in}{3.479321in}}%
\pgfusepath{clip}%
\pgfsetbuttcap%
\pgfsetmiterjoin%
\definecolor{currentfill}{rgb}{0.229806,0.298718,0.753683}%
\pgfsetfillcolor{currentfill}%
\pgfsetlinewidth{1.003750pt}%
\definecolor{currentstroke}{rgb}{0.000000,0.000000,0.000000}%
\pgfsetstrokecolor{currentstroke}%
\pgfsetdash{}{0pt}%
\pgfpathmoveto{\pgfqpoint{3.384241in}{1.437930in}}%
\pgfpathlineto{\pgfqpoint{3.699019in}{1.437930in}}%
\pgfpathlineto{\pgfqpoint{3.699019in}{1.902611in}}%
\pgfpathlineto{\pgfqpoint{3.384241in}{1.902611in}}%
\pgfpathlineto{\pgfqpoint{3.384241in}{1.437930in}}%
\pgfpathclose%
\pgfusepath{stroke,fill}%
\end{pgfscope}%
\begin{pgfscope}%
\pgfpathrectangle{\pgfqpoint{0.603704in}{0.370679in}}{\pgfqpoint{5.246296in}{3.479321in}}%
\pgfusepath{clip}%
\pgfsetbuttcap%
\pgfsetmiterjoin%
\definecolor{currentfill}{rgb}{0.229806,0.298718,0.753683}%
\pgfsetfillcolor{currentfill}%
\pgfsetlinewidth{1.003750pt}%
\definecolor{currentstroke}{rgb}{0.000000,0.000000,0.000000}%
\pgfsetstrokecolor{currentstroke}%
\pgfsetdash{}{0pt}%
\pgfpathmoveto{\pgfqpoint{4.433500in}{1.380629in}}%
\pgfpathlineto{\pgfqpoint{4.748278in}{1.380629in}}%
\pgfpathlineto{\pgfqpoint{4.748278in}{1.825902in}}%
\pgfpathlineto{\pgfqpoint{4.433500in}{1.825902in}}%
\pgfpathlineto{\pgfqpoint{4.433500in}{1.380629in}}%
\pgfpathclose%
\pgfusepath{stroke,fill}%
\end{pgfscope}%
\begin{pgfscope}%
\pgfpathrectangle{\pgfqpoint{0.603704in}{0.370679in}}{\pgfqpoint{5.246296in}{3.479321in}}%
\pgfusepath{clip}%
\pgfsetbuttcap%
\pgfsetmiterjoin%
\definecolor{currentfill}{rgb}{0.705673,0.015556,0.150233}%
\pgfsetfillcolor{currentfill}%
\pgfsetlinewidth{1.003750pt}%
\definecolor{currentstroke}{rgb}{0.000000,0.000000,0.000000}%
\pgfsetstrokecolor{currentstroke}%
\pgfsetdash{}{0pt}%
\pgfpathmoveto{\pgfqpoint{1.705426in}{1.450011in}}%
\pgfpathlineto{\pgfqpoint{2.020204in}{1.450011in}}%
\pgfpathlineto{\pgfqpoint{2.020204in}{1.888473in}}%
\pgfpathlineto{\pgfqpoint{1.705426in}{1.888473in}}%
\pgfpathlineto{\pgfqpoint{1.705426in}{1.450011in}}%
\pgfpathclose%
\pgfusepath{stroke,fill}%
\end{pgfscope}%
\begin{pgfscope}%
\pgfpathrectangle{\pgfqpoint{0.603704in}{0.370679in}}{\pgfqpoint{5.246296in}{3.479321in}}%
\pgfusepath{clip}%
\pgfsetbuttcap%
\pgfsetmiterjoin%
\definecolor{currentfill}{rgb}{0.705673,0.015556,0.150233}%
\pgfsetfillcolor{currentfill}%
\pgfsetlinewidth{1.003750pt}%
\definecolor{currentstroke}{rgb}{0.000000,0.000000,0.000000}%
\pgfsetstrokecolor{currentstroke}%
\pgfsetdash{}{0pt}%
\pgfpathmoveto{\pgfqpoint{2.754685in}{1.353976in}}%
\pgfpathlineto{\pgfqpoint{3.069463in}{1.353976in}}%
\pgfpathlineto{\pgfqpoint{3.069463in}{1.974415in}}%
\pgfpathlineto{\pgfqpoint{2.754685in}{1.974415in}}%
\pgfpathlineto{\pgfqpoint{2.754685in}{1.353976in}}%
\pgfpathclose%
\pgfusepath{stroke,fill}%
\end{pgfscope}%
\begin{pgfscope}%
\pgfpathrectangle{\pgfqpoint{0.603704in}{0.370679in}}{\pgfqpoint{5.246296in}{3.479321in}}%
\pgfusepath{clip}%
\pgfsetbuttcap%
\pgfsetmiterjoin%
\definecolor{currentfill}{rgb}{0.705673,0.015556,0.150233}%
\pgfsetfillcolor{currentfill}%
\pgfsetlinewidth{1.003750pt}%
\definecolor{currentstroke}{rgb}{0.000000,0.000000,0.000000}%
\pgfsetstrokecolor{currentstroke}%
\pgfsetdash{}{0pt}%
\pgfpathmoveto{\pgfqpoint{3.803945in}{1.416760in}}%
\pgfpathlineto{\pgfqpoint{4.118722in}{1.416760in}}%
\pgfpathlineto{\pgfqpoint{4.118722in}{2.197422in}}%
\pgfpathlineto{\pgfqpoint{3.803945in}{2.197422in}}%
\pgfpathlineto{\pgfqpoint{3.803945in}{1.416760in}}%
\pgfpathclose%
\pgfusepath{stroke,fill}%
\end{pgfscope}%
\begin{pgfscope}%
\pgfpathrectangle{\pgfqpoint{0.603704in}{0.370679in}}{\pgfqpoint{5.246296in}{3.479321in}}%
\pgfusepath{clip}%
\pgfsetbuttcap%
\pgfsetmiterjoin%
\definecolor{currentfill}{rgb}{0.705673,0.015556,0.150233}%
\pgfsetfillcolor{currentfill}%
\pgfsetlinewidth{1.003750pt}%
\definecolor{currentstroke}{rgb}{0.000000,0.000000,0.000000}%
\pgfsetstrokecolor{currentstroke}%
\pgfsetdash{}{0pt}%
\pgfpathmoveto{\pgfqpoint{4.853204in}{1.369016in}}%
\pgfpathlineto{\pgfqpoint{5.167982in}{1.369016in}}%
\pgfpathlineto{\pgfqpoint{5.167982in}{1.935875in}}%
\pgfpathlineto{\pgfqpoint{4.853204in}{1.935875in}}%
\pgfpathlineto{\pgfqpoint{4.853204in}{1.369016in}}%
\pgfpathclose%
\pgfusepath{stroke,fill}%
\end{pgfscope}%
\begin{pgfscope}%
\pgfpathrectangle{\pgfqpoint{0.603704in}{0.370679in}}{\pgfqpoint{5.246296in}{3.479321in}}%
\pgfusepath{clip}%
\pgfsetrectcap%
\pgfsetroundjoin%
\pgfsetlinewidth{1.003750pt}%
\definecolor{currentstroke}{rgb}{1.000000,0.498039,0.054902}%
\pgfsetstrokecolor{currentstroke}%
\pgfsetdash{}{0pt}%
\pgfpathmoveto{\pgfqpoint{1.285723in}{1.606103in}}%
\pgfpathlineto{\pgfqpoint{1.600500in}{1.606103in}}%
\pgfusepath{stroke}%
\end{pgfscope}%
\begin{pgfscope}%
\pgfpathrectangle{\pgfqpoint{0.603704in}{0.370679in}}{\pgfqpoint{5.246296in}{3.479321in}}%
\pgfusepath{clip}%
\pgfsetrectcap%
\pgfsetroundjoin%
\pgfsetlinewidth{1.003750pt}%
\definecolor{currentstroke}{rgb}{1.000000,0.498039,0.054902}%
\pgfsetstrokecolor{currentstroke}%
\pgfsetdash{}{0pt}%
\pgfpathmoveto{\pgfqpoint{2.334982in}{1.611480in}}%
\pgfpathlineto{\pgfqpoint{2.649760in}{1.611480in}}%
\pgfusepath{stroke}%
\end{pgfscope}%
\begin{pgfscope}%
\pgfpathrectangle{\pgfqpoint{0.603704in}{0.370679in}}{\pgfqpoint{5.246296in}{3.479321in}}%
\pgfusepath{clip}%
\pgfsetrectcap%
\pgfsetroundjoin%
\pgfsetlinewidth{1.003750pt}%
\definecolor{currentstroke}{rgb}{1.000000,0.498039,0.054902}%
\pgfsetstrokecolor{currentstroke}%
\pgfsetdash{}{0pt}%
\pgfpathmoveto{\pgfqpoint{3.384241in}{1.621948in}}%
\pgfpathlineto{\pgfqpoint{3.699019in}{1.621948in}}%
\pgfusepath{stroke}%
\end{pgfscope}%
\begin{pgfscope}%
\pgfpathrectangle{\pgfqpoint{0.603704in}{0.370679in}}{\pgfqpoint{5.246296in}{3.479321in}}%
\pgfusepath{clip}%
\pgfsetrectcap%
\pgfsetroundjoin%
\pgfsetlinewidth{1.003750pt}%
\definecolor{currentstroke}{rgb}{1.000000,0.498039,0.054902}%
\pgfsetstrokecolor{currentstroke}%
\pgfsetdash{}{0pt}%
\pgfpathmoveto{\pgfqpoint{4.433500in}{1.559088in}}%
\pgfpathlineto{\pgfqpoint{4.748278in}{1.559088in}}%
\pgfusepath{stroke}%
\end{pgfscope}%
\begin{pgfscope}%
\pgfpathrectangle{\pgfqpoint{0.603704in}{0.370679in}}{\pgfqpoint{5.246296in}{3.479321in}}%
\pgfusepath{clip}%
\pgfsetrectcap%
\pgfsetroundjoin%
\pgfsetlinewidth{1.003750pt}%
\definecolor{currentstroke}{rgb}{1.000000,0.498039,0.054902}%
\pgfsetstrokecolor{currentstroke}%
\pgfsetdash{}{0pt}%
\pgfpathmoveto{\pgfqpoint{1.705426in}{1.675771in}}%
\pgfpathlineto{\pgfqpoint{2.020204in}{1.675771in}}%
\pgfusepath{stroke}%
\end{pgfscope}%
\begin{pgfscope}%
\pgfpathrectangle{\pgfqpoint{0.603704in}{0.370679in}}{\pgfqpoint{5.246296in}{3.479321in}}%
\pgfusepath{clip}%
\pgfsetrectcap%
\pgfsetroundjoin%
\pgfsetlinewidth{1.003750pt}%
\definecolor{currentstroke}{rgb}{1.000000,0.498039,0.054902}%
\pgfsetstrokecolor{currentstroke}%
\pgfsetdash{}{0pt}%
\pgfpathmoveto{\pgfqpoint{2.754685in}{1.621500in}}%
\pgfpathlineto{\pgfqpoint{3.069463in}{1.621500in}}%
\pgfusepath{stroke}%
\end{pgfscope}%
\begin{pgfscope}%
\pgfpathrectangle{\pgfqpoint{0.603704in}{0.370679in}}{\pgfqpoint{5.246296in}{3.479321in}}%
\pgfusepath{clip}%
\pgfsetrectcap%
\pgfsetroundjoin%
\pgfsetlinewidth{1.003750pt}%
\definecolor{currentstroke}{rgb}{1.000000,0.498039,0.054902}%
\pgfsetstrokecolor{currentstroke}%
\pgfsetdash{}{0pt}%
\pgfpathmoveto{\pgfqpoint{3.803945in}{1.695197in}}%
\pgfpathlineto{\pgfqpoint{4.118722in}{1.695197in}}%
\pgfusepath{stroke}%
\end{pgfscope}%
\begin{pgfscope}%
\pgfpathrectangle{\pgfqpoint{0.603704in}{0.370679in}}{\pgfqpoint{5.246296in}{3.479321in}}%
\pgfusepath{clip}%
\pgfsetrectcap%
\pgfsetroundjoin%
\pgfsetlinewidth{1.003750pt}%
\definecolor{currentstroke}{rgb}{1.000000,0.498039,0.054902}%
\pgfsetstrokecolor{currentstroke}%
\pgfsetdash{}{0pt}%
\pgfpathmoveto{\pgfqpoint{4.853204in}{1.602681in}}%
\pgfpathlineto{\pgfqpoint{5.167982in}{1.602681in}}%
\pgfusepath{stroke}%
\end{pgfscope}%
\begin{pgfscope}%
\pgfsetrectcap%
\pgfsetmiterjoin%
\pgfsetlinewidth{0.803000pt}%
\definecolor{currentstroke}{rgb}{0.000000,0.000000,0.000000}%
\pgfsetstrokecolor{currentstroke}%
\pgfsetdash{}{0pt}%
\pgfpathmoveto{\pgfqpoint{0.603704in}{0.370679in}}%
\pgfpathlineto{\pgfqpoint{0.603704in}{3.850000in}}%
\pgfusepath{stroke}%
\end{pgfscope}%
\begin{pgfscope}%
\pgfsetrectcap%
\pgfsetmiterjoin%
\pgfsetlinewidth{0.803000pt}%
\definecolor{currentstroke}{rgb}{0.000000,0.000000,0.000000}%
\pgfsetstrokecolor{currentstroke}%
\pgfsetdash{}{0pt}%
\pgfpathmoveto{\pgfqpoint{5.850000in}{0.370679in}}%
\pgfpathlineto{\pgfqpoint{5.850000in}{3.850000in}}%
\pgfusepath{stroke}%
\end{pgfscope}%
\begin{pgfscope}%
\pgfsetrectcap%
\pgfsetmiterjoin%
\pgfsetlinewidth{0.803000pt}%
\definecolor{currentstroke}{rgb}{0.000000,0.000000,0.000000}%
\pgfsetstrokecolor{currentstroke}%
\pgfsetdash{}{0pt}%
\pgfpathmoveto{\pgfqpoint{0.603704in}{0.370679in}}%
\pgfpathlineto{\pgfqpoint{5.850000in}{0.370679in}}%
\pgfusepath{stroke}%
\end{pgfscope}%
\begin{pgfscope}%
\pgfsetrectcap%
\pgfsetmiterjoin%
\pgfsetlinewidth{0.803000pt}%
\definecolor{currentstroke}{rgb}{0.000000,0.000000,0.000000}%
\pgfsetstrokecolor{currentstroke}%
\pgfsetdash{}{0pt}%
\pgfpathmoveto{\pgfqpoint{0.603704in}{3.850000in}}%
\pgfpathlineto{\pgfqpoint{5.850000in}{3.850000in}}%
\pgfusepath{stroke}%
\end{pgfscope}%
\begin{pgfscope}%
\definecolor{textcolor}{rgb}{0.000000,0.000000,0.000000}%
\pgfsetstrokecolor{textcolor}%
\pgfsetfillcolor{textcolor}%
\pgftext[x=1.652963in,y=2.751267in,,bottom]{\color{textcolor}\rmfamily\fontsize{10.000000}{12.000000}\selectfont n.s.}%
\end{pgfscope}%
\begin{pgfscope}%
\definecolor{textcolor}{rgb}{0.000000,0.000000,0.000000}%
\pgfsetstrokecolor{textcolor}%
\pgfsetfillcolor{textcolor}%
\pgftext[x=2.702222in,y=3.056471in,,bottom]{\color{textcolor}\rmfamily\fontsize{10.000000}{12.000000}\selectfont n.s.}%
\end{pgfscope}%
\begin{pgfscope}%
\definecolor{textcolor}{rgb}{0.000000,0.000000,0.000000}%
\pgfsetstrokecolor{textcolor}%
\pgfsetfillcolor{textcolor}%
\pgftext[x=3.751482in,y=3.544796in,,bottom]{\color{textcolor}\rmfamily\fontsize{10.000000}{12.000000}\selectfont \(\displaystyle p\) \(\displaystyle <\) 0.05}%
\end{pgfscope}%
\begin{pgfscope}%
\definecolor{textcolor}{rgb}{0.000000,0.000000,0.000000}%
\pgfsetstrokecolor{textcolor}%
\pgfsetfillcolor{textcolor}%
\pgftext[x=4.800741in,y=2.934389in,,bottom]{\color{textcolor}\rmfamily\fontsize{10.000000}{12.000000}\selectfont n.s.}%
\end{pgfscope}%
\begin{pgfscope}%
\pgfsetbuttcap%
\pgfsetmiterjoin%
\definecolor{currentfill}{rgb}{1.000000,1.000000,1.000000}%
\pgfsetfillcolor{currentfill}%
\pgfsetfillopacity{0.800000}%
\pgfsetlinewidth{1.003750pt}%
\definecolor{currentstroke}{rgb}{0.800000,0.800000,0.800000}%
\pgfsetstrokecolor{currentstroke}%
\pgfsetstrokeopacity{0.800000}%
\pgfsetdash{}{0pt}%
\pgfpathmoveto{\pgfqpoint{4.819521in}{3.351543in}}%
\pgfpathlineto{\pgfqpoint{5.752778in}{3.351543in}}%
\pgfpathquadraticcurveto{\pgfqpoint{5.780556in}{3.351543in}}{\pgfqpoint{5.780556in}{3.379321in}}%
\pgfpathlineto{\pgfqpoint{5.780556in}{3.752778in}}%
\pgfpathquadraticcurveto{\pgfqpoint{5.780556in}{3.780556in}}{\pgfqpoint{5.752778in}{3.780556in}}%
\pgfpathlineto{\pgfqpoint{4.819521in}{3.780556in}}%
\pgfpathquadraticcurveto{\pgfqpoint{4.791743in}{3.780556in}}{\pgfqpoint{4.791743in}{3.752778in}}%
\pgfpathlineto{\pgfqpoint{4.791743in}{3.379321in}}%
\pgfpathquadraticcurveto{\pgfqpoint{4.791743in}{3.351543in}}{\pgfqpoint{4.819521in}{3.351543in}}%
\pgfpathclose%
\pgfusepath{stroke,fill}%
\end{pgfscope}%
\begin{pgfscope}%
\pgfsetrectcap%
\pgfsetroundjoin%
\pgfsetlinewidth{1.505625pt}%
\definecolor{currentstroke}{rgb}{0.229806,0.298718,0.753683}%
\pgfsetstrokecolor{currentstroke}%
\pgfsetdash{}{0pt}%
\pgfpathmoveto{\pgfqpoint{4.847299in}{3.676389in}}%
\pgfpathlineto{\pgfqpoint{5.125076in}{3.676389in}}%
\pgfusepath{stroke}%
\end{pgfscope}%
\begin{pgfscope}%
\definecolor{textcolor}{rgb}{0.000000,0.000000,0.000000}%
\pgfsetstrokecolor{textcolor}%
\pgfsetfillcolor{textcolor}%
\pgftext[x=5.236187in,y=3.627778in,left,base]{\color{textcolor}\rmfamily\fontsize{10.000000}{12.000000}\selectfont At-Risk}%
\end{pgfscope}%
\begin{pgfscope}%
\pgfsetrectcap%
\pgfsetroundjoin%
\pgfsetlinewidth{1.505625pt}%
\definecolor{currentstroke}{rgb}{0.705673,0.015556,0.150233}%
\pgfsetstrokecolor{currentstroke}%
\pgfsetdash{}{0pt}%
\pgfpathmoveto{\pgfqpoint{4.847299in}{3.482716in}}%
\pgfpathlineto{\pgfqpoint{5.125076in}{3.482716in}}%
\pgfusepath{stroke}%
\end{pgfscope}%
\begin{pgfscope}%
\definecolor{textcolor}{rgb}{0.000000,0.000000,0.000000}%
\pgfsetstrokecolor{textcolor}%
\pgfsetfillcolor{textcolor}%
\pgftext[x=5.236187in,y=3.434105in,left,base]{\color{textcolor}\rmfamily\fontsize{10.000000}{12.000000}\selectfont No-Risk}%
\end{pgfscope}%
\end{pgfpicture}%
\makeatother%
\endgroup%

%% file: LaTeX/images/quadrant_boxplot_info.pgf
\begingroup%
\makeatletter%
\begin{pgfpicture}%
\pgfpathrectangle{\pgfpointorigin}{\pgfqpoint{6.000000in}{4.000000in}}%
\pgfusepath{use as bounding box, clip}%
\begin{pgfscope}%
\pgfsetbuttcap%
\pgfsetmiterjoin%
\pgfsetlinewidth{0.000000pt}%
\definecolor{currentstroke}{rgb}{1.000000,1.000000,1.000000}%
\pgfsetstrokecolor{currentstroke}%
\pgfsetstrokeopacity{0.000000}%
\pgfsetdash{}{0pt}%
\pgfpathmoveto{\pgfqpoint{0.000000in}{0.000000in}}%
\pgfpathlineto{\pgfqpoint{6.000000in}{0.000000in}}%
\pgfpathlineto{\pgfqpoint{6.000000in}{4.000000in}}%
\pgfpathlineto{\pgfqpoint{0.000000in}{4.000000in}}%
\pgfpathclose%
\pgfusepath{}%
\end{pgfscope}%
\begin{pgfscope}%
\pgfsetbuttcap%
\pgfsetmiterjoin%
\definecolor{currentfill}{rgb}{1.000000,1.000000,1.000000}%
\pgfsetfillcolor{currentfill}%
\pgfsetlinewidth{0.000000pt}%
\definecolor{currentstroke}{rgb}{0.000000,0.000000,0.000000}%
\pgfsetstrokecolor{currentstroke}%
\pgfsetstrokeopacity{0.000000}%
\pgfsetdash{}{0pt}%
\pgfpathmoveto{\pgfqpoint{0.634568in}{0.370679in}}%
\pgfpathlineto{\pgfqpoint{5.850000in}{0.370679in}}%
\pgfpathlineto{\pgfqpoint{5.850000in}{3.801775in}}%
\pgfpathlineto{\pgfqpoint{0.634568in}{3.801775in}}%
\pgfpathclose%
\pgfusepath{fill}%
\end{pgfscope}%
\begin{pgfscope}%
\pgfsetbuttcap%
\pgfsetroundjoin%
\definecolor{currentfill}{rgb}{0.000000,0.000000,0.000000}%
\pgfsetfillcolor{currentfill}%
\pgfsetlinewidth{0.803000pt}%
\definecolor{currentstroke}{rgb}{0.000000,0.000000,0.000000}%
\pgfsetstrokecolor{currentstroke}%
\pgfsetdash{}{0pt}%
\pgfsys@defobject{currentmarker}{\pgfqpoint{0.000000in}{-0.048611in}}{\pgfqpoint{0.000000in}{0.000000in}}{%
\pgfpathmoveto{\pgfqpoint{0.000000in}{0.000000in}}%
\pgfpathlineto{\pgfqpoint{0.000000in}{-0.048611in}}%
\pgfusepath{stroke,fill}%
}%
\begin{pgfscope}%
\pgfsys@transformshift{1.677655in}{0.370679in}%
\pgfsys@useobject{currentmarker}{}%
\end{pgfscope}%
\end{pgfscope}%
\begin{pgfscope}%
\definecolor{textcolor}{rgb}{0.000000,0.000000,0.000000}%
\pgfsetstrokecolor{textcolor}%
\pgfsetfillcolor{textcolor}%
\pgftext[x=1.677655in,y=0.273457in,,top]{\color{textcolor}\rmfamily\fontsize{10.000000}{12.000000}\selectfont Q1: Happiness}%
\end{pgfscope}%
\begin{pgfscope}%
\pgfsetbuttcap%
\pgfsetroundjoin%
\definecolor{currentfill}{rgb}{0.000000,0.000000,0.000000}%
\pgfsetfillcolor{currentfill}%
\pgfsetlinewidth{0.803000pt}%
\definecolor{currentstroke}{rgb}{0.000000,0.000000,0.000000}%
\pgfsetstrokecolor{currentstroke}%
\pgfsetdash{}{0pt}%
\pgfsys@defobject{currentmarker}{\pgfqpoint{0.000000in}{-0.048611in}}{\pgfqpoint{0.000000in}{0.000000in}}{%
\pgfpathmoveto{\pgfqpoint{0.000000in}{0.000000in}}%
\pgfpathlineto{\pgfqpoint{0.000000in}{-0.048611in}}%
\pgfusepath{stroke,fill}%
}%
\begin{pgfscope}%
\pgfsys@transformshift{2.720741in}{0.370679in}%
\pgfsys@useobject{currentmarker}{}%
\end{pgfscope}%
\end{pgfscope}%
\begin{pgfscope}%
\definecolor{textcolor}{rgb}{0.000000,0.000000,0.000000}%
\pgfsetstrokecolor{textcolor}%
\pgfsetfillcolor{textcolor}%
\pgftext[x=2.720741in,y=0.273457in,,top]{\color{textcolor}\rmfamily\fontsize{10.000000}{12.000000}\selectfont Q2: Anger}%
\end{pgfscope}%
\begin{pgfscope}%
\pgfsetbuttcap%
\pgfsetroundjoin%
\definecolor{currentfill}{rgb}{0.000000,0.000000,0.000000}%
\pgfsetfillcolor{currentfill}%
\pgfsetlinewidth{0.803000pt}%
\definecolor{currentstroke}{rgb}{0.000000,0.000000,0.000000}%
\pgfsetstrokecolor{currentstroke}%
\pgfsetdash{}{0pt}%
\pgfsys@defobject{currentmarker}{\pgfqpoint{0.000000in}{-0.048611in}}{\pgfqpoint{0.000000in}{0.000000in}}{%
\pgfpathmoveto{\pgfqpoint{0.000000in}{0.000000in}}%
\pgfpathlineto{\pgfqpoint{0.000000in}{-0.048611in}}%
\pgfusepath{stroke,fill}%
}%
\begin{pgfscope}%
\pgfsys@transformshift{3.763827in}{0.370679in}%
\pgfsys@useobject{currentmarker}{}%
\end{pgfscope}%
\end{pgfscope}%
\begin{pgfscope}%
\definecolor{textcolor}{rgb}{0.000000,0.000000,0.000000}%
\pgfsetstrokecolor{textcolor}%
\pgfsetfillcolor{textcolor}%
\pgftext[x=3.763827in,y=0.273457in,,top]{\color{textcolor}\rmfamily\fontsize{10.000000}{12.000000}\selectfont Q3: Sad}%
\end{pgfscope}%
\begin{pgfscope}%
\pgfsetbuttcap%
\pgfsetroundjoin%
\definecolor{currentfill}{rgb}{0.000000,0.000000,0.000000}%
\pgfsetfillcolor{currentfill}%
\pgfsetlinewidth{0.803000pt}%
\definecolor{currentstroke}{rgb}{0.000000,0.000000,0.000000}%
\pgfsetstrokecolor{currentstroke}%
\pgfsetdash{}{0pt}%
\pgfsys@defobject{currentmarker}{\pgfqpoint{0.000000in}{-0.048611in}}{\pgfqpoint{0.000000in}{0.000000in}}{%
\pgfpathmoveto{\pgfqpoint{0.000000in}{0.000000in}}%
\pgfpathlineto{\pgfqpoint{0.000000in}{-0.048611in}}%
\pgfusepath{stroke,fill}%
}%
\begin{pgfscope}%
\pgfsys@transformshift{4.806914in}{0.370679in}%
\pgfsys@useobject{currentmarker}{}%
\end{pgfscope}%
\end{pgfscope}%
\begin{pgfscope}%
\definecolor{textcolor}{rgb}{0.000000,0.000000,0.000000}%
\pgfsetstrokecolor{textcolor}%
\pgfsetfillcolor{textcolor}%
\pgftext[x=4.806914in,y=0.273457in,,top]{\color{textcolor}\rmfamily\fontsize{10.000000}{12.000000}\selectfont Q4: Tenderness}%
\end{pgfscope}%
\begin{pgfscope}%
\pgfsetbuttcap%
\pgfsetroundjoin%
\definecolor{currentfill}{rgb}{0.000000,0.000000,0.000000}%
\pgfsetfillcolor{currentfill}%
\pgfsetlinewidth{0.803000pt}%
\definecolor{currentstroke}{rgb}{0.000000,0.000000,0.000000}%
\pgfsetstrokecolor{currentstroke}%
\pgfsetdash{}{0pt}%
\pgfsys@defobject{currentmarker}{\pgfqpoint{-0.048611in}{0.000000in}}{\pgfqpoint{-0.000000in}{0.000000in}}{%
\pgfpathmoveto{\pgfqpoint{-0.000000in}{0.000000in}}%
\pgfpathlineto{\pgfqpoint{-0.048611in}{0.000000in}}%
\pgfusepath{stroke,fill}%
}%
\begin{pgfscope}%
\pgfsys@transformshift{0.634568in}{0.370679in}%
\pgfsys@useobject{currentmarker}{}%
\end{pgfscope}%
\end{pgfscope}%
\begin{pgfscope}%
\definecolor{textcolor}{rgb}{0.000000,0.000000,0.000000}%
\pgfsetstrokecolor{textcolor}%
\pgfsetfillcolor{textcolor}%
\pgftext[x=0.467902in, y=0.322454in, left, base]{\color{textcolor}\rmfamily\fontsize{10.000000}{12.000000}\selectfont \(\displaystyle {0}\)}%
\end{pgfscope}%
\begin{pgfscope}%
\pgfsetbuttcap%
\pgfsetroundjoin%
\definecolor{currentfill}{rgb}{0.000000,0.000000,0.000000}%
\pgfsetfillcolor{currentfill}%
\pgfsetlinewidth{0.803000pt}%
\definecolor{currentstroke}{rgb}{0.000000,0.000000,0.000000}%
\pgfsetstrokecolor{currentstroke}%
\pgfsetdash{}{0pt}%
\pgfsys@defobject{currentmarker}{\pgfqpoint{-0.048611in}{0.000000in}}{\pgfqpoint{-0.000000in}{0.000000in}}{%
\pgfpathmoveto{\pgfqpoint{-0.000000in}{0.000000in}}%
\pgfpathlineto{\pgfqpoint{-0.048611in}{0.000000in}}%
\pgfusepath{stroke,fill}%
}%
\begin{pgfscope}%
\pgfsys@transformshift{0.634568in}{0.860835in}%
\pgfsys@useobject{currentmarker}{}%
\end{pgfscope}%
\end{pgfscope}%
\begin{pgfscope}%
\definecolor{textcolor}{rgb}{0.000000,0.000000,0.000000}%
\pgfsetstrokecolor{textcolor}%
\pgfsetfillcolor{textcolor}%
\pgftext[x=0.398457in, y=0.812610in, left, base]{\color{textcolor}\rmfamily\fontsize{10.000000}{12.000000}\selectfont \(\displaystyle {50}\)}%
\end{pgfscope}%
\begin{pgfscope}%
\pgfsetbuttcap%
\pgfsetroundjoin%
\definecolor{currentfill}{rgb}{0.000000,0.000000,0.000000}%
\pgfsetfillcolor{currentfill}%
\pgfsetlinewidth{0.803000pt}%
\definecolor{currentstroke}{rgb}{0.000000,0.000000,0.000000}%
\pgfsetstrokecolor{currentstroke}%
\pgfsetdash{}{0pt}%
\pgfsys@defobject{currentmarker}{\pgfqpoint{-0.048611in}{0.000000in}}{\pgfqpoint{-0.000000in}{0.000000in}}{%
\pgfpathmoveto{\pgfqpoint{-0.000000in}{0.000000in}}%
\pgfpathlineto{\pgfqpoint{-0.048611in}{0.000000in}}%
\pgfusepath{stroke,fill}%
}%
\begin{pgfscope}%
\pgfsys@transformshift{0.634568in}{1.350992in}%
\pgfsys@useobject{currentmarker}{}%
\end{pgfscope}%
\end{pgfscope}%
\begin{pgfscope}%
\definecolor{textcolor}{rgb}{0.000000,0.000000,0.000000}%
\pgfsetstrokecolor{textcolor}%
\pgfsetfillcolor{textcolor}%
\pgftext[x=0.329012in, y=1.302767in, left, base]{\color{textcolor}\rmfamily\fontsize{10.000000}{12.000000}\selectfont \(\displaystyle {100}\)}%
\end{pgfscope}%
\begin{pgfscope}%
\pgfsetbuttcap%
\pgfsetroundjoin%
\definecolor{currentfill}{rgb}{0.000000,0.000000,0.000000}%
\pgfsetfillcolor{currentfill}%
\pgfsetlinewidth{0.803000pt}%
\definecolor{currentstroke}{rgb}{0.000000,0.000000,0.000000}%
\pgfsetstrokecolor{currentstroke}%
\pgfsetdash{}{0pt}%
\pgfsys@defobject{currentmarker}{\pgfqpoint{-0.048611in}{0.000000in}}{\pgfqpoint{-0.000000in}{0.000000in}}{%
\pgfpathmoveto{\pgfqpoint{-0.000000in}{0.000000in}}%
\pgfpathlineto{\pgfqpoint{-0.048611in}{0.000000in}}%
\pgfusepath{stroke,fill}%
}%
\begin{pgfscope}%
\pgfsys@transformshift{0.634568in}{1.841149in}%
\pgfsys@useobject{currentmarker}{}%
\end{pgfscope}%
\end{pgfscope}%
\begin{pgfscope}%
\definecolor{textcolor}{rgb}{0.000000,0.000000,0.000000}%
\pgfsetstrokecolor{textcolor}%
\pgfsetfillcolor{textcolor}%
\pgftext[x=0.329012in, y=1.792923in, left, base]{\color{textcolor}\rmfamily\fontsize{10.000000}{12.000000}\selectfont \(\displaystyle {150}\)}%
\end{pgfscope}%
\begin{pgfscope}%
\pgfsetbuttcap%
\pgfsetroundjoin%
\definecolor{currentfill}{rgb}{0.000000,0.000000,0.000000}%
\pgfsetfillcolor{currentfill}%
\pgfsetlinewidth{0.803000pt}%
\definecolor{currentstroke}{rgb}{0.000000,0.000000,0.000000}%
\pgfsetstrokecolor{currentstroke}%
\pgfsetdash{}{0pt}%
\pgfsys@defobject{currentmarker}{\pgfqpoint{-0.048611in}{0.000000in}}{\pgfqpoint{-0.000000in}{0.000000in}}{%
\pgfpathmoveto{\pgfqpoint{-0.000000in}{0.000000in}}%
\pgfpathlineto{\pgfqpoint{-0.048611in}{0.000000in}}%
\pgfusepath{stroke,fill}%
}%
\begin{pgfscope}%
\pgfsys@transformshift{0.634568in}{2.331305in}%
\pgfsys@useobject{currentmarker}{}%
\end{pgfscope}%
\end{pgfscope}%
\begin{pgfscope}%
\definecolor{textcolor}{rgb}{0.000000,0.000000,0.000000}%
\pgfsetstrokecolor{textcolor}%
\pgfsetfillcolor{textcolor}%
\pgftext[x=0.329012in, y=2.283080in, left, base]{\color{textcolor}\rmfamily\fontsize{10.000000}{12.000000}\selectfont \(\displaystyle {200}\)}%
\end{pgfscope}%
\begin{pgfscope}%
\pgfsetbuttcap%
\pgfsetroundjoin%
\definecolor{currentfill}{rgb}{0.000000,0.000000,0.000000}%
\pgfsetfillcolor{currentfill}%
\pgfsetlinewidth{0.803000pt}%
\definecolor{currentstroke}{rgb}{0.000000,0.000000,0.000000}%
\pgfsetstrokecolor{currentstroke}%
\pgfsetdash{}{0pt}%
\pgfsys@defobject{currentmarker}{\pgfqpoint{-0.048611in}{0.000000in}}{\pgfqpoint{-0.000000in}{0.000000in}}{%
\pgfpathmoveto{\pgfqpoint{-0.000000in}{0.000000in}}%
\pgfpathlineto{\pgfqpoint{-0.048611in}{0.000000in}}%
\pgfusepath{stroke,fill}%
}%
\begin{pgfscope}%
\pgfsys@transformshift{0.634568in}{2.821462in}%
\pgfsys@useobject{currentmarker}{}%
\end{pgfscope}%
\end{pgfscope}%
\begin{pgfscope}%
\definecolor{textcolor}{rgb}{0.000000,0.000000,0.000000}%
\pgfsetstrokecolor{textcolor}%
\pgfsetfillcolor{textcolor}%
\pgftext[x=0.329012in, y=2.773236in, left, base]{\color{textcolor}\rmfamily\fontsize{10.000000}{12.000000}\selectfont \(\displaystyle {250}\)}%
\end{pgfscope}%
\begin{pgfscope}%
\pgfsetbuttcap%
\pgfsetroundjoin%
\definecolor{currentfill}{rgb}{0.000000,0.000000,0.000000}%
\pgfsetfillcolor{currentfill}%
\pgfsetlinewidth{0.803000pt}%
\definecolor{currentstroke}{rgb}{0.000000,0.000000,0.000000}%
\pgfsetstrokecolor{currentstroke}%
\pgfsetdash{}{0pt}%
\pgfsys@defobject{currentmarker}{\pgfqpoint{-0.048611in}{0.000000in}}{\pgfqpoint{-0.000000in}{0.000000in}}{%
\pgfpathmoveto{\pgfqpoint{-0.000000in}{0.000000in}}%
\pgfpathlineto{\pgfqpoint{-0.048611in}{0.000000in}}%
\pgfusepath{stroke,fill}%
}%
\begin{pgfscope}%
\pgfsys@transformshift{0.634568in}{3.311618in}%
\pgfsys@useobject{currentmarker}{}%
\end{pgfscope}%
\end{pgfscope}%
\begin{pgfscope}%
\definecolor{textcolor}{rgb}{0.000000,0.000000,0.000000}%
\pgfsetstrokecolor{textcolor}%
\pgfsetfillcolor{textcolor}%
\pgftext[x=0.329012in, y=3.263393in, left, base]{\color{textcolor}\rmfamily\fontsize{10.000000}{12.000000}\selectfont \(\displaystyle {300}\)}%
\end{pgfscope}%
\begin{pgfscope}%
\pgfsetbuttcap%
\pgfsetroundjoin%
\definecolor{currentfill}{rgb}{0.000000,0.000000,0.000000}%
\pgfsetfillcolor{currentfill}%
\pgfsetlinewidth{0.803000pt}%
\definecolor{currentstroke}{rgb}{0.000000,0.000000,0.000000}%
\pgfsetstrokecolor{currentstroke}%
\pgfsetdash{}{0pt}%
\pgfsys@defobject{currentmarker}{\pgfqpoint{-0.048611in}{0.000000in}}{\pgfqpoint{-0.000000in}{0.000000in}}{%
\pgfpathmoveto{\pgfqpoint{-0.000000in}{0.000000in}}%
\pgfpathlineto{\pgfqpoint{-0.048611in}{0.000000in}}%
\pgfusepath{stroke,fill}%
}%
\begin{pgfscope}%
\pgfsys@transformshift{0.634568in}{3.801775in}%
\pgfsys@useobject{currentmarker}{}%
\end{pgfscope}%
\end{pgfscope}%
\begin{pgfscope}%
\definecolor{textcolor}{rgb}{0.000000,0.000000,0.000000}%
\pgfsetstrokecolor{textcolor}%
\pgfsetfillcolor{textcolor}%
\pgftext[x=0.329012in, y=3.753549in, left, base]{\color{textcolor}\rmfamily\fontsize{10.000000}{12.000000}\selectfont \(\displaystyle {350}\)}%
\end{pgfscope}%
\begin{pgfscope}%
\definecolor{textcolor}{rgb}{0.000000,0.000000,0.000000}%
\pgfsetstrokecolor{textcolor}%
\pgfsetfillcolor{textcolor}%
\pgftext[x=0.273457in,y=2.086227in,,bottom,rotate=90.000000]{\color{textcolor}\rmfamily\fontsize{10.000000}{12.000000}\selectfont AIC}%
\end{pgfscope}%
\begin{pgfscope}%
\pgfpathrectangle{\pgfqpoint{0.634568in}{0.370679in}}{\pgfqpoint{5.215432in}{3.431096in}}%
\pgfusepath{clip}%
\pgfsetrectcap%
\pgfsetroundjoin%
\pgfsetlinewidth{1.003750pt}%
\definecolor{currentstroke}{rgb}{0.000000,0.000000,0.000000}%
\pgfsetstrokecolor{currentstroke}%
\pgfsetdash{}{0pt}%
\pgfpathmoveto{\pgfqpoint{1.469038in}{1.743787in}}%
\pgfpathlineto{\pgfqpoint{1.469038in}{1.289370in}}%
\pgfusepath{stroke}%
\end{pgfscope}%
\begin{pgfscope}%
\pgfpathrectangle{\pgfqpoint{0.634568in}{0.370679in}}{\pgfqpoint{5.215432in}{3.431096in}}%
\pgfusepath{clip}%
\pgfsetrectcap%
\pgfsetroundjoin%
\pgfsetlinewidth{1.003750pt}%
\definecolor{currentstroke}{rgb}{0.000000,0.000000,0.000000}%
\pgfsetstrokecolor{currentstroke}%
\pgfsetdash{}{0pt}%
\pgfpathmoveto{\pgfqpoint{1.469038in}{2.327175in}}%
\pgfpathlineto{\pgfqpoint{1.469038in}{3.184270in}}%
\pgfusepath{stroke}%
\end{pgfscope}%
\begin{pgfscope}%
\pgfpathrectangle{\pgfqpoint{0.634568in}{0.370679in}}{\pgfqpoint{5.215432in}{3.431096in}}%
\pgfusepath{clip}%
\pgfsetrectcap%
\pgfsetroundjoin%
\pgfsetlinewidth{1.003750pt}%
\definecolor{currentstroke}{rgb}{0.000000,0.000000,0.000000}%
\pgfsetstrokecolor{currentstroke}%
\pgfsetdash{}{0pt}%
\pgfpathmoveto{\pgfqpoint{1.390806in}{1.289370in}}%
\pgfpathlineto{\pgfqpoint{1.547269in}{1.289370in}}%
\pgfusepath{stroke}%
\end{pgfscope}%
\begin{pgfscope}%
\pgfpathrectangle{\pgfqpoint{0.634568in}{0.370679in}}{\pgfqpoint{5.215432in}{3.431096in}}%
\pgfusepath{clip}%
\pgfsetrectcap%
\pgfsetroundjoin%
\pgfsetlinewidth{1.003750pt}%
\definecolor{currentstroke}{rgb}{0.000000,0.000000,0.000000}%
\pgfsetstrokecolor{currentstroke}%
\pgfsetdash{}{0pt}%
\pgfpathmoveto{\pgfqpoint{1.390806in}{3.184270in}}%
\pgfpathlineto{\pgfqpoint{1.547269in}{3.184270in}}%
\pgfusepath{stroke}%
\end{pgfscope}%
\begin{pgfscope}%
\pgfpathrectangle{\pgfqpoint{0.634568in}{0.370679in}}{\pgfqpoint{5.215432in}{3.431096in}}%
\pgfusepath{clip}%
\pgfsetrectcap%
\pgfsetroundjoin%
\pgfsetlinewidth{1.003750pt}%
\definecolor{currentstroke}{rgb}{0.000000,0.000000,0.000000}%
\pgfsetstrokecolor{currentstroke}%
\pgfsetdash{}{0pt}%
\pgfpathmoveto{\pgfqpoint{2.512124in}{1.463903in}}%
\pgfpathlineto{\pgfqpoint{2.512124in}{0.852064in}}%
\pgfusepath{stroke}%
\end{pgfscope}%
\begin{pgfscope}%
\pgfpathrectangle{\pgfqpoint{0.634568in}{0.370679in}}{\pgfqpoint{5.215432in}{3.431096in}}%
\pgfusepath{clip}%
\pgfsetrectcap%
\pgfsetroundjoin%
\pgfsetlinewidth{1.003750pt}%
\definecolor{currentstroke}{rgb}{0.000000,0.000000,0.000000}%
\pgfsetstrokecolor{currentstroke}%
\pgfsetdash{}{0pt}%
\pgfpathmoveto{\pgfqpoint{2.512124in}{1.960334in}}%
\pgfpathlineto{\pgfqpoint{2.512124in}{2.692461in}}%
\pgfusepath{stroke}%
\end{pgfscope}%
\begin{pgfscope}%
\pgfpathrectangle{\pgfqpoint{0.634568in}{0.370679in}}{\pgfqpoint{5.215432in}{3.431096in}}%
\pgfusepath{clip}%
\pgfsetrectcap%
\pgfsetroundjoin%
\pgfsetlinewidth{1.003750pt}%
\definecolor{currentstroke}{rgb}{0.000000,0.000000,0.000000}%
\pgfsetstrokecolor{currentstroke}%
\pgfsetdash{}{0pt}%
\pgfpathmoveto{\pgfqpoint{2.433892in}{0.852064in}}%
\pgfpathlineto{\pgfqpoint{2.590355in}{0.852064in}}%
\pgfusepath{stroke}%
\end{pgfscope}%
\begin{pgfscope}%
\pgfpathrectangle{\pgfqpoint{0.634568in}{0.370679in}}{\pgfqpoint{5.215432in}{3.431096in}}%
\pgfusepath{clip}%
\pgfsetrectcap%
\pgfsetroundjoin%
\pgfsetlinewidth{1.003750pt}%
\definecolor{currentstroke}{rgb}{0.000000,0.000000,0.000000}%
\pgfsetstrokecolor{currentstroke}%
\pgfsetdash{}{0pt}%
\pgfpathmoveto{\pgfqpoint{2.433892in}{2.692461in}}%
\pgfpathlineto{\pgfqpoint{2.590355in}{2.692461in}}%
\pgfusepath{stroke}%
\end{pgfscope}%
\begin{pgfscope}%
\pgfpathrectangle{\pgfqpoint{0.634568in}{0.370679in}}{\pgfqpoint{5.215432in}{3.431096in}}%
\pgfusepath{clip}%
\pgfsetrectcap%
\pgfsetroundjoin%
\pgfsetlinewidth{1.003750pt}%
\definecolor{currentstroke}{rgb}{0.000000,0.000000,0.000000}%
\pgfsetstrokecolor{currentstroke}%
\pgfsetdash{}{0pt}%
\pgfpathmoveto{\pgfqpoint{3.555210in}{1.238344in}}%
\pgfpathlineto{\pgfqpoint{3.555210in}{0.591327in}}%
\pgfusepath{stroke}%
\end{pgfscope}%
\begin{pgfscope}%
\pgfpathrectangle{\pgfqpoint{0.634568in}{0.370679in}}{\pgfqpoint{5.215432in}{3.431096in}}%
\pgfusepath{clip}%
\pgfsetrectcap%
\pgfsetroundjoin%
\pgfsetlinewidth{1.003750pt}%
\definecolor{currentstroke}{rgb}{0.000000,0.000000,0.000000}%
\pgfsetstrokecolor{currentstroke}%
\pgfsetdash{}{0pt}%
\pgfpathmoveto{\pgfqpoint{3.555210in}{1.699219in}}%
\pgfpathlineto{\pgfqpoint{3.555210in}{2.360935in}}%
\pgfusepath{stroke}%
\end{pgfscope}%
\begin{pgfscope}%
\pgfpathrectangle{\pgfqpoint{0.634568in}{0.370679in}}{\pgfqpoint{5.215432in}{3.431096in}}%
\pgfusepath{clip}%
\pgfsetrectcap%
\pgfsetroundjoin%
\pgfsetlinewidth{1.003750pt}%
\definecolor{currentstroke}{rgb}{0.000000,0.000000,0.000000}%
\pgfsetstrokecolor{currentstroke}%
\pgfsetdash{}{0pt}%
\pgfpathmoveto{\pgfqpoint{3.476979in}{0.591327in}}%
\pgfpathlineto{\pgfqpoint{3.633442in}{0.591327in}}%
\pgfusepath{stroke}%
\end{pgfscope}%
\begin{pgfscope}%
\pgfpathrectangle{\pgfqpoint{0.634568in}{0.370679in}}{\pgfqpoint{5.215432in}{3.431096in}}%
\pgfusepath{clip}%
\pgfsetrectcap%
\pgfsetroundjoin%
\pgfsetlinewidth{1.003750pt}%
\definecolor{currentstroke}{rgb}{0.000000,0.000000,0.000000}%
\pgfsetstrokecolor{currentstroke}%
\pgfsetdash{}{0pt}%
\pgfpathmoveto{\pgfqpoint{3.476979in}{2.360935in}}%
\pgfpathlineto{\pgfqpoint{3.633442in}{2.360935in}}%
\pgfusepath{stroke}%
\end{pgfscope}%
\begin{pgfscope}%
\pgfpathrectangle{\pgfqpoint{0.634568in}{0.370679in}}{\pgfqpoint{5.215432in}{3.431096in}}%
\pgfusepath{clip}%
\pgfsetrectcap%
\pgfsetroundjoin%
\pgfsetlinewidth{1.003750pt}%
\definecolor{currentstroke}{rgb}{0.000000,0.000000,0.000000}%
\pgfsetstrokecolor{currentstroke}%
\pgfsetdash{}{0pt}%
\pgfpathmoveto{\pgfqpoint{4.598296in}{1.648789in}}%
\pgfpathlineto{\pgfqpoint{4.598296in}{1.032068in}}%
\pgfusepath{stroke}%
\end{pgfscope}%
\begin{pgfscope}%
\pgfpathrectangle{\pgfqpoint{0.634568in}{0.370679in}}{\pgfqpoint{5.215432in}{3.431096in}}%
\pgfusepath{clip}%
\pgfsetrectcap%
\pgfsetroundjoin%
\pgfsetlinewidth{1.003750pt}%
\definecolor{currentstroke}{rgb}{0.000000,0.000000,0.000000}%
\pgfsetstrokecolor{currentstroke}%
\pgfsetdash{}{0pt}%
\pgfpathmoveto{\pgfqpoint{4.598296in}{2.060850in}}%
\pgfpathlineto{\pgfqpoint{4.598296in}{2.671224in}}%
\pgfusepath{stroke}%
\end{pgfscope}%
\begin{pgfscope}%
\pgfpathrectangle{\pgfqpoint{0.634568in}{0.370679in}}{\pgfqpoint{5.215432in}{3.431096in}}%
\pgfusepath{clip}%
\pgfsetrectcap%
\pgfsetroundjoin%
\pgfsetlinewidth{1.003750pt}%
\definecolor{currentstroke}{rgb}{0.000000,0.000000,0.000000}%
\pgfsetstrokecolor{currentstroke}%
\pgfsetdash{}{0pt}%
\pgfpathmoveto{\pgfqpoint{4.520065in}{1.032068in}}%
\pgfpathlineto{\pgfqpoint{4.676528in}{1.032068in}}%
\pgfusepath{stroke}%
\end{pgfscope}%
\begin{pgfscope}%
\pgfpathrectangle{\pgfqpoint{0.634568in}{0.370679in}}{\pgfqpoint{5.215432in}{3.431096in}}%
\pgfusepath{clip}%
\pgfsetrectcap%
\pgfsetroundjoin%
\pgfsetlinewidth{1.003750pt}%
\definecolor{currentstroke}{rgb}{0.000000,0.000000,0.000000}%
\pgfsetstrokecolor{currentstroke}%
\pgfsetdash{}{0pt}%
\pgfpathmoveto{\pgfqpoint{4.520065in}{2.671224in}}%
\pgfpathlineto{\pgfqpoint{4.676528in}{2.671224in}}%
\pgfusepath{stroke}%
\end{pgfscope}%
\begin{pgfscope}%
\pgfpathrectangle{\pgfqpoint{0.634568in}{0.370679in}}{\pgfqpoint{5.215432in}{3.431096in}}%
\pgfusepath{clip}%
\pgfsetrectcap%
\pgfsetroundjoin%
\pgfsetlinewidth{1.003750pt}%
\definecolor{currentstroke}{rgb}{0.000000,0.000000,0.000000}%
\pgfsetstrokecolor{currentstroke}%
\pgfsetdash{}{0pt}%
\pgfpathmoveto{\pgfqpoint{1.886272in}{1.721399in}}%
\pgfpathlineto{\pgfqpoint{1.886272in}{1.128874in}}%
\pgfusepath{stroke}%
\end{pgfscope}%
\begin{pgfscope}%
\pgfpathrectangle{\pgfqpoint{0.634568in}{0.370679in}}{\pgfqpoint{5.215432in}{3.431096in}}%
\pgfusepath{clip}%
\pgfsetrectcap%
\pgfsetroundjoin%
\pgfsetlinewidth{1.003750pt}%
\definecolor{currentstroke}{rgb}{0.000000,0.000000,0.000000}%
\pgfsetstrokecolor{currentstroke}%
\pgfsetdash{}{0pt}%
\pgfpathmoveto{\pgfqpoint{1.886272in}{2.162089in}}%
\pgfpathlineto{\pgfqpoint{1.886272in}{2.729576in}}%
\pgfusepath{stroke}%
\end{pgfscope}%
\begin{pgfscope}%
\pgfpathrectangle{\pgfqpoint{0.634568in}{0.370679in}}{\pgfqpoint{5.215432in}{3.431096in}}%
\pgfusepath{clip}%
\pgfsetrectcap%
\pgfsetroundjoin%
\pgfsetlinewidth{1.003750pt}%
\definecolor{currentstroke}{rgb}{0.000000,0.000000,0.000000}%
\pgfsetstrokecolor{currentstroke}%
\pgfsetdash{}{0pt}%
\pgfpathmoveto{\pgfqpoint{1.808041in}{1.128874in}}%
\pgfpathlineto{\pgfqpoint{1.964504in}{1.128874in}}%
\pgfusepath{stroke}%
\end{pgfscope}%
\begin{pgfscope}%
\pgfpathrectangle{\pgfqpoint{0.634568in}{0.370679in}}{\pgfqpoint{5.215432in}{3.431096in}}%
\pgfusepath{clip}%
\pgfsetrectcap%
\pgfsetroundjoin%
\pgfsetlinewidth{1.003750pt}%
\definecolor{currentstroke}{rgb}{0.000000,0.000000,0.000000}%
\pgfsetstrokecolor{currentstroke}%
\pgfsetdash{}{0pt}%
\pgfpathmoveto{\pgfqpoint{1.808041in}{2.729576in}}%
\pgfpathlineto{\pgfqpoint{1.964504in}{2.729576in}}%
\pgfusepath{stroke}%
\end{pgfscope}%
\begin{pgfscope}%
\pgfpathrectangle{\pgfqpoint{0.634568in}{0.370679in}}{\pgfqpoint{5.215432in}{3.431096in}}%
\pgfusepath{clip}%
\pgfsetrectcap%
\pgfsetroundjoin%
\pgfsetlinewidth{1.003750pt}%
\definecolor{currentstroke}{rgb}{0.000000,0.000000,0.000000}%
\pgfsetstrokecolor{currentstroke}%
\pgfsetdash{}{0pt}%
\pgfpathmoveto{\pgfqpoint{2.929358in}{1.413062in}}%
\pgfpathlineto{\pgfqpoint{2.929358in}{0.732711in}}%
\pgfusepath{stroke}%
\end{pgfscope}%
\begin{pgfscope}%
\pgfpathrectangle{\pgfqpoint{0.634568in}{0.370679in}}{\pgfqpoint{5.215432in}{3.431096in}}%
\pgfusepath{clip}%
\pgfsetrectcap%
\pgfsetroundjoin%
\pgfsetlinewidth{1.003750pt}%
\definecolor{currentstroke}{rgb}{0.000000,0.000000,0.000000}%
\pgfsetstrokecolor{currentstroke}%
\pgfsetdash{}{0pt}%
\pgfpathmoveto{\pgfqpoint{2.929358in}{1.929975in}}%
\pgfpathlineto{\pgfqpoint{2.929358in}{2.614720in}}%
\pgfusepath{stroke}%
\end{pgfscope}%
\begin{pgfscope}%
\pgfpathrectangle{\pgfqpoint{0.634568in}{0.370679in}}{\pgfqpoint{5.215432in}{3.431096in}}%
\pgfusepath{clip}%
\pgfsetrectcap%
\pgfsetroundjoin%
\pgfsetlinewidth{1.003750pt}%
\definecolor{currentstroke}{rgb}{0.000000,0.000000,0.000000}%
\pgfsetstrokecolor{currentstroke}%
\pgfsetdash{}{0pt}%
\pgfpathmoveto{\pgfqpoint{2.851127in}{0.732711in}}%
\pgfpathlineto{\pgfqpoint{3.007590in}{0.732711in}}%
\pgfusepath{stroke}%
\end{pgfscope}%
\begin{pgfscope}%
\pgfpathrectangle{\pgfqpoint{0.634568in}{0.370679in}}{\pgfqpoint{5.215432in}{3.431096in}}%
\pgfusepath{clip}%
\pgfsetrectcap%
\pgfsetroundjoin%
\pgfsetlinewidth{1.003750pt}%
\definecolor{currentstroke}{rgb}{0.000000,0.000000,0.000000}%
\pgfsetstrokecolor{currentstroke}%
\pgfsetdash{}{0pt}%
\pgfpathmoveto{\pgfqpoint{2.851127in}{2.614720in}}%
\pgfpathlineto{\pgfqpoint{3.007590in}{2.614720in}}%
\pgfusepath{stroke}%
\end{pgfscope}%
\begin{pgfscope}%
\pgfpathrectangle{\pgfqpoint{0.634568in}{0.370679in}}{\pgfqpoint{5.215432in}{3.431096in}}%
\pgfusepath{clip}%
\pgfsetrectcap%
\pgfsetroundjoin%
\pgfsetlinewidth{1.003750pt}%
\definecolor{currentstroke}{rgb}{0.000000,0.000000,0.000000}%
\pgfsetstrokecolor{currentstroke}%
\pgfsetdash{}{0pt}%
\pgfpathmoveto{\pgfqpoint{3.972445in}{1.171988in}}%
\pgfpathlineto{\pgfqpoint{3.972445in}{0.450413in}}%
\pgfusepath{stroke}%
\end{pgfscope}%
\begin{pgfscope}%
\pgfpathrectangle{\pgfqpoint{0.634568in}{0.370679in}}{\pgfqpoint{5.215432in}{3.431096in}}%
\pgfusepath{clip}%
\pgfsetrectcap%
\pgfsetroundjoin%
\pgfsetlinewidth{1.003750pt}%
\definecolor{currentstroke}{rgb}{0.000000,0.000000,0.000000}%
\pgfsetstrokecolor{currentstroke}%
\pgfsetdash{}{0pt}%
\pgfpathmoveto{\pgfqpoint{3.972445in}{1.677848in}}%
\pgfpathlineto{\pgfqpoint{3.972445in}{2.293306in}}%
\pgfusepath{stroke}%
\end{pgfscope}%
\begin{pgfscope}%
\pgfpathrectangle{\pgfqpoint{0.634568in}{0.370679in}}{\pgfqpoint{5.215432in}{3.431096in}}%
\pgfusepath{clip}%
\pgfsetrectcap%
\pgfsetroundjoin%
\pgfsetlinewidth{1.003750pt}%
\definecolor{currentstroke}{rgb}{0.000000,0.000000,0.000000}%
\pgfsetstrokecolor{currentstroke}%
\pgfsetdash{}{0pt}%
\pgfpathmoveto{\pgfqpoint{3.894213in}{0.450413in}}%
\pgfpathlineto{\pgfqpoint{4.050676in}{0.450413in}}%
\pgfusepath{stroke}%
\end{pgfscope}%
\begin{pgfscope}%
\pgfpathrectangle{\pgfqpoint{0.634568in}{0.370679in}}{\pgfqpoint{5.215432in}{3.431096in}}%
\pgfusepath{clip}%
\pgfsetrectcap%
\pgfsetroundjoin%
\pgfsetlinewidth{1.003750pt}%
\definecolor{currentstroke}{rgb}{0.000000,0.000000,0.000000}%
\pgfsetstrokecolor{currentstroke}%
\pgfsetdash{}{0pt}%
\pgfpathmoveto{\pgfqpoint{3.894213in}{2.293306in}}%
\pgfpathlineto{\pgfqpoint{4.050676in}{2.293306in}}%
\pgfusepath{stroke}%
\end{pgfscope}%
\begin{pgfscope}%
\pgfpathrectangle{\pgfqpoint{0.634568in}{0.370679in}}{\pgfqpoint{5.215432in}{3.431096in}}%
\pgfusepath{clip}%
\pgfsetrectcap%
\pgfsetroundjoin%
\pgfsetlinewidth{1.003750pt}%
\definecolor{currentstroke}{rgb}{0.000000,0.000000,0.000000}%
\pgfsetstrokecolor{currentstroke}%
\pgfsetdash{}{0pt}%
\pgfpathmoveto{\pgfqpoint{5.015531in}{1.537477in}}%
\pgfpathlineto{\pgfqpoint{5.015531in}{0.878942in}}%
\pgfusepath{stroke}%
\end{pgfscope}%
\begin{pgfscope}%
\pgfpathrectangle{\pgfqpoint{0.634568in}{0.370679in}}{\pgfqpoint{5.215432in}{3.431096in}}%
\pgfusepath{clip}%
\pgfsetrectcap%
\pgfsetroundjoin%
\pgfsetlinewidth{1.003750pt}%
\definecolor{currentstroke}{rgb}{0.000000,0.000000,0.000000}%
\pgfsetstrokecolor{currentstroke}%
\pgfsetdash{}{0pt}%
\pgfpathmoveto{\pgfqpoint{5.015531in}{1.985629in}}%
\pgfpathlineto{\pgfqpoint{5.015531in}{2.569426in}}%
\pgfusepath{stroke}%
\end{pgfscope}%
\begin{pgfscope}%
\pgfpathrectangle{\pgfqpoint{0.634568in}{0.370679in}}{\pgfqpoint{5.215432in}{3.431096in}}%
\pgfusepath{clip}%
\pgfsetrectcap%
\pgfsetroundjoin%
\pgfsetlinewidth{1.003750pt}%
\definecolor{currentstroke}{rgb}{0.000000,0.000000,0.000000}%
\pgfsetstrokecolor{currentstroke}%
\pgfsetdash{}{0pt}%
\pgfpathmoveto{\pgfqpoint{4.937299in}{0.878942in}}%
\pgfpathlineto{\pgfqpoint{5.093762in}{0.878942in}}%
\pgfusepath{stroke}%
\end{pgfscope}%
\begin{pgfscope}%
\pgfpathrectangle{\pgfqpoint{0.634568in}{0.370679in}}{\pgfqpoint{5.215432in}{3.431096in}}%
\pgfusepath{clip}%
\pgfsetrectcap%
\pgfsetroundjoin%
\pgfsetlinewidth{1.003750pt}%
\definecolor{currentstroke}{rgb}{0.000000,0.000000,0.000000}%
\pgfsetstrokecolor{currentstroke}%
\pgfsetdash{}{0pt}%
\pgfpathmoveto{\pgfqpoint{4.937299in}{2.569426in}}%
\pgfpathlineto{\pgfqpoint{5.093762in}{2.569426in}}%
\pgfusepath{stroke}%
\end{pgfscope}%
\begin{pgfscope}%
\pgfpathrectangle{\pgfqpoint{0.634568in}{0.370679in}}{\pgfqpoint{5.215432in}{3.431096in}}%
\pgfusepath{clip}%
\pgfsetrectcap%
\pgfsetroundjoin%
\pgfsetlinewidth{1.505625pt}%
\definecolor{currentstroke}{rgb}{0.000000,0.000000,0.000000}%
\pgfsetstrokecolor{currentstroke}%
\pgfsetdash{}{0pt}%
\pgfpathmoveto{\pgfqpoint{1.469038in}{3.311618in}}%
\pgfpathlineto{\pgfqpoint{1.469038in}{3.409649in}}%
\pgfpathlineto{\pgfqpoint{1.886272in}{3.409649in}}%
\pgfpathlineto{\pgfqpoint{1.886272in}{3.311618in}}%
\pgfusepath{stroke}%
\end{pgfscope}%
\begin{pgfscope}%
\pgfpathrectangle{\pgfqpoint{0.634568in}{0.370679in}}{\pgfqpoint{5.215432in}{3.431096in}}%
\pgfusepath{clip}%
\pgfsetrectcap%
\pgfsetroundjoin%
\pgfsetlinewidth{1.505625pt}%
\definecolor{currentstroke}{rgb}{0.000000,0.000000,0.000000}%
\pgfsetstrokecolor{currentstroke}%
\pgfsetdash{}{0pt}%
\pgfpathmoveto{\pgfqpoint{2.512124in}{2.821462in}}%
\pgfpathlineto{\pgfqpoint{2.512124in}{2.919493in}}%
\pgfpathlineto{\pgfqpoint{2.929358in}{2.919493in}}%
\pgfpathlineto{\pgfqpoint{2.929358in}{2.821462in}}%
\pgfusepath{stroke}%
\end{pgfscope}%
\begin{pgfscope}%
\pgfpathrectangle{\pgfqpoint{0.634568in}{0.370679in}}{\pgfqpoint{5.215432in}{3.431096in}}%
\pgfusepath{clip}%
\pgfsetrectcap%
\pgfsetroundjoin%
\pgfsetlinewidth{1.505625pt}%
\definecolor{currentstroke}{rgb}{0.000000,0.000000,0.000000}%
\pgfsetstrokecolor{currentstroke}%
\pgfsetdash{}{0pt}%
\pgfpathmoveto{\pgfqpoint{3.555210in}{2.429336in}}%
\pgfpathlineto{\pgfqpoint{3.555210in}{2.527368in}}%
\pgfpathlineto{\pgfqpoint{3.972445in}{2.527368in}}%
\pgfpathlineto{\pgfqpoint{3.972445in}{2.429336in}}%
\pgfusepath{stroke}%
\end{pgfscope}%
\begin{pgfscope}%
\pgfpathrectangle{\pgfqpoint{0.634568in}{0.370679in}}{\pgfqpoint{5.215432in}{3.431096in}}%
\pgfusepath{clip}%
\pgfsetrectcap%
\pgfsetroundjoin%
\pgfsetlinewidth{1.505625pt}%
\definecolor{currentstroke}{rgb}{0.000000,0.000000,0.000000}%
\pgfsetstrokecolor{currentstroke}%
\pgfsetdash{}{0pt}%
\pgfpathmoveto{\pgfqpoint{4.598296in}{2.772446in}}%
\pgfpathlineto{\pgfqpoint{4.598296in}{2.870477in}}%
\pgfpathlineto{\pgfqpoint{5.015531in}{2.870477in}}%
\pgfpathlineto{\pgfqpoint{5.015531in}{2.772446in}}%
\pgfusepath{stroke}%
\end{pgfscope}%
\begin{pgfscope}%
\pgfpathrectangle{\pgfqpoint{0.634568in}{0.370679in}}{\pgfqpoint{5.215432in}{3.431096in}}%
\pgfusepath{clip}%
\pgfsetbuttcap%
\pgfsetmiterjoin%
\definecolor{currentfill}{rgb}{0.229806,0.298718,0.753683}%
\pgfsetfillcolor{currentfill}%
\pgfsetlinewidth{1.003750pt}%
\definecolor{currentstroke}{rgb}{0.000000,0.000000,0.000000}%
\pgfsetstrokecolor{currentstroke}%
\pgfsetdash{}{0pt}%
\pgfpathmoveto{\pgfqpoint{1.312575in}{1.743787in}}%
\pgfpathlineto{\pgfqpoint{1.625500in}{1.743787in}}%
\pgfpathlineto{\pgfqpoint{1.625500in}{2.327175in}}%
\pgfpathlineto{\pgfqpoint{1.312575in}{2.327175in}}%
\pgfpathlineto{\pgfqpoint{1.312575in}{1.743787in}}%
\pgfpathclose%
\pgfusepath{stroke,fill}%
\end{pgfscope}%
\begin{pgfscope}%
\pgfpathrectangle{\pgfqpoint{0.634568in}{0.370679in}}{\pgfqpoint{5.215432in}{3.431096in}}%
\pgfusepath{clip}%
\pgfsetbuttcap%
\pgfsetmiterjoin%
\definecolor{currentfill}{rgb}{0.229806,0.298718,0.753683}%
\pgfsetfillcolor{currentfill}%
\pgfsetlinewidth{1.003750pt}%
\definecolor{currentstroke}{rgb}{0.000000,0.000000,0.000000}%
\pgfsetstrokecolor{currentstroke}%
\pgfsetdash{}{0pt}%
\pgfpathmoveto{\pgfqpoint{2.355661in}{1.463903in}}%
\pgfpathlineto{\pgfqpoint{2.668587in}{1.463903in}}%
\pgfpathlineto{\pgfqpoint{2.668587in}{1.960334in}}%
\pgfpathlineto{\pgfqpoint{2.355661in}{1.960334in}}%
\pgfpathlineto{\pgfqpoint{2.355661in}{1.463903in}}%
\pgfpathclose%
\pgfusepath{stroke,fill}%
\end{pgfscope}%
\begin{pgfscope}%
\pgfpathrectangle{\pgfqpoint{0.634568in}{0.370679in}}{\pgfqpoint{5.215432in}{3.431096in}}%
\pgfusepath{clip}%
\pgfsetbuttcap%
\pgfsetmiterjoin%
\definecolor{currentfill}{rgb}{0.229806,0.298718,0.753683}%
\pgfsetfillcolor{currentfill}%
\pgfsetlinewidth{1.003750pt}%
\definecolor{currentstroke}{rgb}{0.000000,0.000000,0.000000}%
\pgfsetstrokecolor{currentstroke}%
\pgfsetdash{}{0pt}%
\pgfpathmoveto{\pgfqpoint{3.398747in}{1.238344in}}%
\pgfpathlineto{\pgfqpoint{3.711673in}{1.238344in}}%
\pgfpathlineto{\pgfqpoint{3.711673in}{1.699219in}}%
\pgfpathlineto{\pgfqpoint{3.398747in}{1.699219in}}%
\pgfpathlineto{\pgfqpoint{3.398747in}{1.238344in}}%
\pgfpathclose%
\pgfusepath{stroke,fill}%
\end{pgfscope}%
\begin{pgfscope}%
\pgfpathrectangle{\pgfqpoint{0.634568in}{0.370679in}}{\pgfqpoint{5.215432in}{3.431096in}}%
\pgfusepath{clip}%
\pgfsetbuttcap%
\pgfsetmiterjoin%
\definecolor{currentfill}{rgb}{0.229806,0.298718,0.753683}%
\pgfsetfillcolor{currentfill}%
\pgfsetlinewidth{1.003750pt}%
\definecolor{currentstroke}{rgb}{0.000000,0.000000,0.000000}%
\pgfsetstrokecolor{currentstroke}%
\pgfsetdash{}{0pt}%
\pgfpathmoveto{\pgfqpoint{4.441833in}{1.648789in}}%
\pgfpathlineto{\pgfqpoint{4.754759in}{1.648789in}}%
\pgfpathlineto{\pgfqpoint{4.754759in}{2.060850in}}%
\pgfpathlineto{\pgfqpoint{4.441833in}{2.060850in}}%
\pgfpathlineto{\pgfqpoint{4.441833in}{1.648789in}}%
\pgfpathclose%
\pgfusepath{stroke,fill}%
\end{pgfscope}%
\begin{pgfscope}%
\pgfpathrectangle{\pgfqpoint{0.634568in}{0.370679in}}{\pgfqpoint{5.215432in}{3.431096in}}%
\pgfusepath{clip}%
\pgfsetbuttcap%
\pgfsetmiterjoin%
\definecolor{currentfill}{rgb}{0.705673,0.015556,0.150233}%
\pgfsetfillcolor{currentfill}%
\pgfsetlinewidth{1.003750pt}%
\definecolor{currentstroke}{rgb}{0.000000,0.000000,0.000000}%
\pgfsetstrokecolor{currentstroke}%
\pgfsetdash{}{0pt}%
\pgfpathmoveto{\pgfqpoint{1.729809in}{1.721399in}}%
\pgfpathlineto{\pgfqpoint{2.042735in}{1.721399in}}%
\pgfpathlineto{\pgfqpoint{2.042735in}{2.162089in}}%
\pgfpathlineto{\pgfqpoint{1.729809in}{2.162089in}}%
\pgfpathlineto{\pgfqpoint{1.729809in}{1.721399in}}%
\pgfpathclose%
\pgfusepath{stroke,fill}%
\end{pgfscope}%
\begin{pgfscope}%
\pgfpathrectangle{\pgfqpoint{0.634568in}{0.370679in}}{\pgfqpoint{5.215432in}{3.431096in}}%
\pgfusepath{clip}%
\pgfsetbuttcap%
\pgfsetmiterjoin%
\definecolor{currentfill}{rgb}{0.705673,0.015556,0.150233}%
\pgfsetfillcolor{currentfill}%
\pgfsetlinewidth{1.003750pt}%
\definecolor{currentstroke}{rgb}{0.000000,0.000000,0.000000}%
\pgfsetstrokecolor{currentstroke}%
\pgfsetdash{}{0pt}%
\pgfpathmoveto{\pgfqpoint{2.772895in}{1.413062in}}%
\pgfpathlineto{\pgfqpoint{3.085821in}{1.413062in}}%
\pgfpathlineto{\pgfqpoint{3.085821in}{1.929975in}}%
\pgfpathlineto{\pgfqpoint{2.772895in}{1.929975in}}%
\pgfpathlineto{\pgfqpoint{2.772895in}{1.413062in}}%
\pgfpathclose%
\pgfusepath{stroke,fill}%
\end{pgfscope}%
\begin{pgfscope}%
\pgfpathrectangle{\pgfqpoint{0.634568in}{0.370679in}}{\pgfqpoint{5.215432in}{3.431096in}}%
\pgfusepath{clip}%
\pgfsetbuttcap%
\pgfsetmiterjoin%
\definecolor{currentfill}{rgb}{0.705673,0.015556,0.150233}%
\pgfsetfillcolor{currentfill}%
\pgfsetlinewidth{1.003750pt}%
\definecolor{currentstroke}{rgb}{0.000000,0.000000,0.000000}%
\pgfsetstrokecolor{currentstroke}%
\pgfsetdash{}{0pt}%
\pgfpathmoveto{\pgfqpoint{3.815982in}{1.171988in}}%
\pgfpathlineto{\pgfqpoint{4.128908in}{1.171988in}}%
\pgfpathlineto{\pgfqpoint{4.128908in}{1.677848in}}%
\pgfpathlineto{\pgfqpoint{3.815982in}{1.677848in}}%
\pgfpathlineto{\pgfqpoint{3.815982in}{1.171988in}}%
\pgfpathclose%
\pgfusepath{stroke,fill}%
\end{pgfscope}%
\begin{pgfscope}%
\pgfpathrectangle{\pgfqpoint{0.634568in}{0.370679in}}{\pgfqpoint{5.215432in}{3.431096in}}%
\pgfusepath{clip}%
\pgfsetbuttcap%
\pgfsetmiterjoin%
\definecolor{currentfill}{rgb}{0.705673,0.015556,0.150233}%
\pgfsetfillcolor{currentfill}%
\pgfsetlinewidth{1.003750pt}%
\definecolor{currentstroke}{rgb}{0.000000,0.000000,0.000000}%
\pgfsetstrokecolor{currentstroke}%
\pgfsetdash{}{0pt}%
\pgfpathmoveto{\pgfqpoint{4.859068in}{1.537477in}}%
\pgfpathlineto{\pgfqpoint{5.171994in}{1.537477in}}%
\pgfpathlineto{\pgfqpoint{5.171994in}{1.985629in}}%
\pgfpathlineto{\pgfqpoint{4.859068in}{1.985629in}}%
\pgfpathlineto{\pgfqpoint{4.859068in}{1.537477in}}%
\pgfpathclose%
\pgfusepath{stroke,fill}%
\end{pgfscope}%
\begin{pgfscope}%
\pgfpathrectangle{\pgfqpoint{0.634568in}{0.370679in}}{\pgfqpoint{5.215432in}{3.431096in}}%
\pgfusepath{clip}%
\pgfsetrectcap%
\pgfsetroundjoin%
\pgfsetlinewidth{1.003750pt}%
\definecolor{currentstroke}{rgb}{1.000000,0.498039,0.054902}%
\pgfsetstrokecolor{currentstroke}%
\pgfsetdash{}{0pt}%
\pgfpathmoveto{\pgfqpoint{1.312575in}{1.980362in}}%
\pgfpathlineto{\pgfqpoint{1.625500in}{1.980362in}}%
\pgfusepath{stroke}%
\end{pgfscope}%
\begin{pgfscope}%
\pgfpathrectangle{\pgfqpoint{0.634568in}{0.370679in}}{\pgfqpoint{5.215432in}{3.431096in}}%
\pgfusepath{clip}%
\pgfsetrectcap%
\pgfsetroundjoin%
\pgfsetlinewidth{1.003750pt}%
\definecolor{currentstroke}{rgb}{1.000000,0.498039,0.054902}%
\pgfsetstrokecolor{currentstroke}%
\pgfsetdash{}{0pt}%
\pgfpathmoveto{\pgfqpoint{2.355661in}{1.738591in}}%
\pgfpathlineto{\pgfqpoint{2.668587in}{1.738591in}}%
\pgfusepath{stroke}%
\end{pgfscope}%
\begin{pgfscope}%
\pgfpathrectangle{\pgfqpoint{0.634568in}{0.370679in}}{\pgfqpoint{5.215432in}{3.431096in}}%
\pgfusepath{clip}%
\pgfsetrectcap%
\pgfsetroundjoin%
\pgfsetlinewidth{1.003750pt}%
\definecolor{currentstroke}{rgb}{1.000000,0.498039,0.054902}%
\pgfsetstrokecolor{currentstroke}%
\pgfsetdash{}{0pt}%
\pgfpathmoveto{\pgfqpoint{3.398747in}{1.495647in}}%
\pgfpathlineto{\pgfqpoint{3.711673in}{1.495647in}}%
\pgfusepath{stroke}%
\end{pgfscope}%
\begin{pgfscope}%
\pgfpathrectangle{\pgfqpoint{0.634568in}{0.370679in}}{\pgfqpoint{5.215432in}{3.431096in}}%
\pgfusepath{clip}%
\pgfsetrectcap%
\pgfsetroundjoin%
\pgfsetlinewidth{1.003750pt}%
\definecolor{currentstroke}{rgb}{1.000000,0.498039,0.054902}%
\pgfsetstrokecolor{currentstroke}%
\pgfsetdash{}{0pt}%
\pgfpathmoveto{\pgfqpoint{4.441833in}{1.808479in}}%
\pgfpathlineto{\pgfqpoint{4.754759in}{1.808479in}}%
\pgfusepath{stroke}%
\end{pgfscope}%
\begin{pgfscope}%
\pgfpathrectangle{\pgfqpoint{0.634568in}{0.370679in}}{\pgfqpoint{5.215432in}{3.431096in}}%
\pgfusepath{clip}%
\pgfsetrectcap%
\pgfsetroundjoin%
\pgfsetlinewidth{1.003750pt}%
\definecolor{currentstroke}{rgb}{1.000000,0.498039,0.054902}%
\pgfsetstrokecolor{currentstroke}%
\pgfsetdash{}{0pt}%
\pgfpathmoveto{\pgfqpoint{1.729809in}{1.903045in}}%
\pgfpathlineto{\pgfqpoint{2.042735in}{1.903045in}}%
\pgfusepath{stroke}%
\end{pgfscope}%
\begin{pgfscope}%
\pgfpathrectangle{\pgfqpoint{0.634568in}{0.370679in}}{\pgfqpoint{5.215432in}{3.431096in}}%
\pgfusepath{clip}%
\pgfsetrectcap%
\pgfsetroundjoin%
\pgfsetlinewidth{1.003750pt}%
\definecolor{currentstroke}{rgb}{1.000000,0.498039,0.054902}%
\pgfsetstrokecolor{currentstroke}%
\pgfsetdash{}{0pt}%
\pgfpathmoveto{\pgfqpoint{2.772895in}{1.636092in}}%
\pgfpathlineto{\pgfqpoint{3.085821in}{1.636092in}}%
\pgfusepath{stroke}%
\end{pgfscope}%
\begin{pgfscope}%
\pgfpathrectangle{\pgfqpoint{0.634568in}{0.370679in}}{\pgfqpoint{5.215432in}{3.431096in}}%
\pgfusepath{clip}%
\pgfsetrectcap%
\pgfsetroundjoin%
\pgfsetlinewidth{1.003750pt}%
\definecolor{currentstroke}{rgb}{1.000000,0.498039,0.054902}%
\pgfsetstrokecolor{currentstroke}%
\pgfsetdash{}{0pt}%
\pgfpathmoveto{\pgfqpoint{3.815982in}{1.426226in}}%
\pgfpathlineto{\pgfqpoint{4.128908in}{1.426226in}}%
\pgfusepath{stroke}%
\end{pgfscope}%
\begin{pgfscope}%
\pgfpathrectangle{\pgfqpoint{0.634568in}{0.370679in}}{\pgfqpoint{5.215432in}{3.431096in}}%
\pgfusepath{clip}%
\pgfsetrectcap%
\pgfsetroundjoin%
\pgfsetlinewidth{1.003750pt}%
\definecolor{currentstroke}{rgb}{1.000000,0.498039,0.054902}%
\pgfsetstrokecolor{currentstroke}%
\pgfsetdash{}{0pt}%
\pgfpathmoveto{\pgfqpoint{4.859068in}{1.734409in}}%
\pgfpathlineto{\pgfqpoint{5.171994in}{1.734409in}}%
\pgfusepath{stroke}%
\end{pgfscope}%
\begin{pgfscope}%
\pgfsetrectcap%
\pgfsetmiterjoin%
\pgfsetlinewidth{0.803000pt}%
\definecolor{currentstroke}{rgb}{0.000000,0.000000,0.000000}%
\pgfsetstrokecolor{currentstroke}%
\pgfsetdash{}{0pt}%
\pgfpathmoveto{\pgfqpoint{0.634568in}{0.370679in}}%
\pgfpathlineto{\pgfqpoint{0.634568in}{3.801775in}}%
\pgfusepath{stroke}%
\end{pgfscope}%
\begin{pgfscope}%
\pgfsetrectcap%
\pgfsetmiterjoin%
\pgfsetlinewidth{0.803000pt}%
\definecolor{currentstroke}{rgb}{0.000000,0.000000,0.000000}%
\pgfsetstrokecolor{currentstroke}%
\pgfsetdash{}{0pt}%
\pgfpathmoveto{\pgfqpoint{5.850000in}{0.370679in}}%
\pgfpathlineto{\pgfqpoint{5.850000in}{3.801775in}}%
\pgfusepath{stroke}%
\end{pgfscope}%
\begin{pgfscope}%
\pgfsetrectcap%
\pgfsetmiterjoin%
\pgfsetlinewidth{0.803000pt}%
\definecolor{currentstroke}{rgb}{0.000000,0.000000,0.000000}%
\pgfsetstrokecolor{currentstroke}%
\pgfsetdash{}{0pt}%
\pgfpathmoveto{\pgfqpoint{0.634568in}{0.370679in}}%
\pgfpathlineto{\pgfqpoint{5.850000in}{0.370679in}}%
\pgfusepath{stroke}%
\end{pgfscope}%
\begin{pgfscope}%
\pgfsetrectcap%
\pgfsetmiterjoin%
\pgfsetlinewidth{0.803000pt}%
\definecolor{currentstroke}{rgb}{0.000000,0.000000,0.000000}%
\pgfsetstrokecolor{currentstroke}%
\pgfsetdash{}{0pt}%
\pgfpathmoveto{\pgfqpoint{0.634568in}{3.801775in}}%
\pgfpathlineto{\pgfqpoint{5.850000in}{3.801775in}}%
\pgfusepath{stroke}%
\end{pgfscope}%
\begin{pgfscope}%
\definecolor{textcolor}{rgb}{0.000000,0.000000,0.000000}%
\pgfsetstrokecolor{textcolor}%
\pgfsetfillcolor{textcolor}%
\pgftext[x=1.677655in,y=3.507681in,,bottom]{\color{textcolor}\rmfamily\fontsize{10.000000}{12.000000}\selectfont \(\displaystyle p < 0.05\)}%
\end{pgfscope}%
\begin{pgfscope}%
\definecolor{textcolor}{rgb}{0.000000,0.000000,0.000000}%
\pgfsetstrokecolor{textcolor}%
\pgfsetfillcolor{textcolor}%
\pgftext[x=2.720741in,y=3.017524in,,bottom]{\color{textcolor}\rmfamily\fontsize{10.000000}{12.000000}\selectfont n.s.}%
\end{pgfscope}%
\begin{pgfscope}%
\definecolor{textcolor}{rgb}{0.000000,0.000000,0.000000}%
\pgfsetstrokecolor{textcolor}%
\pgfsetfillcolor{textcolor}%
\pgftext[x=3.763827in,y=2.625399in,,bottom]{\color{textcolor}\rmfamily\fontsize{10.000000}{12.000000}\selectfont \(\displaystyle p = 0.07\)}%
\end{pgfscope}%
\begin{pgfscope}%
\definecolor{textcolor}{rgb}{0.000000,0.000000,0.000000}%
\pgfsetstrokecolor{textcolor}%
\pgfsetfillcolor{textcolor}%
\pgftext[x=4.806914in,y=2.968509in,,bottom]{\color{textcolor}\rmfamily\fontsize{10.000000}{12.000000}\selectfont \(\displaystyle p < 0.01\)}%
\end{pgfscope}%
\begin{pgfscope}%
\pgfsetbuttcap%
\pgfsetmiterjoin%
\definecolor{currentfill}{rgb}{1.000000,1.000000,1.000000}%
\pgfsetfillcolor{currentfill}%
\pgfsetfillopacity{0.800000}%
\pgfsetlinewidth{1.003750pt}%
\definecolor{currentstroke}{rgb}{0.800000,0.800000,0.800000}%
\pgfsetstrokecolor{currentstroke}%
\pgfsetstrokeopacity{0.800000}%
\pgfsetdash{}{0pt}%
\pgfpathmoveto{\pgfqpoint{4.819521in}{3.303318in}}%
\pgfpathlineto{\pgfqpoint{5.752778in}{3.303318in}}%
\pgfpathquadraticcurveto{\pgfqpoint{5.780556in}{3.303318in}}{\pgfqpoint{5.780556in}{3.331096in}}%
\pgfpathlineto{\pgfqpoint{5.780556in}{3.704552in}}%
\pgfpathquadraticcurveto{\pgfqpoint{5.780556in}{3.732330in}}{\pgfqpoint{5.752778in}{3.732330in}}%
\pgfpathlineto{\pgfqpoint{4.819521in}{3.732330in}}%
\pgfpathquadraticcurveto{\pgfqpoint{4.791743in}{3.732330in}}{\pgfqpoint{4.791743in}{3.704552in}}%
\pgfpathlineto{\pgfqpoint{4.791743in}{3.331096in}}%
\pgfpathquadraticcurveto{\pgfqpoint{4.791743in}{3.303318in}}{\pgfqpoint{4.819521in}{3.303318in}}%
\pgfpathclose%
\pgfusepath{stroke,fill}%
\end{pgfscope}%
\begin{pgfscope}%
\pgfsetrectcap%
\pgfsetroundjoin%
\pgfsetlinewidth{1.505625pt}%
\definecolor{currentstroke}{rgb}{0.229806,0.298718,0.753683}%
\pgfsetstrokecolor{currentstroke}%
\pgfsetdash{}{0pt}%
\pgfpathmoveto{\pgfqpoint{4.847299in}{3.628164in}}%
\pgfpathlineto{\pgfqpoint{5.125076in}{3.628164in}}%
\pgfusepath{stroke}%
\end{pgfscope}%
\begin{pgfscope}%
\definecolor{textcolor}{rgb}{0.000000,0.000000,0.000000}%
\pgfsetstrokecolor{textcolor}%
\pgfsetfillcolor{textcolor}%
\pgftext[x=5.236187in,y=3.579553in,left,base]{\color{textcolor}\rmfamily\fontsize{10.000000}{12.000000}\selectfont At-Risk}%
\end{pgfscope}%
\begin{pgfscope}%
\pgfsetrectcap%
\pgfsetroundjoin%
\pgfsetlinewidth{1.505625pt}%
\definecolor{currentstroke}{rgb}{0.705673,0.015556,0.150233}%
\pgfsetstrokecolor{currentstroke}%
\pgfsetdash{}{0pt}%
\pgfpathmoveto{\pgfqpoint{4.847299in}{3.434491in}}%
\pgfpathlineto{\pgfqpoint{5.125076in}{3.434491in}}%
\pgfusepath{stroke}%
\end{pgfscope}%
\begin{pgfscope}%
\definecolor{textcolor}{rgb}{0.000000,0.000000,0.000000}%
\pgfsetstrokecolor{textcolor}%
\pgfsetfillcolor{textcolor}%
\pgftext[x=5.236187in,y=3.385880in,left,base]{\color{textcolor}\rmfamily\fontsize{10.000000}{12.000000}\selectfont No-Risk}%
\end{pgfscope}%
\end{pgfpicture}%
\makeatother%
\endgroup%

%% file: LaTeX/discussion.tex
This is the first study, to the best of our knowledge, to consider lyrical features in relation individuals at risk for depression on online music streaming platforms. Our results confirmed most of our hypotheses, in addition to revealing novel insights.


\subsection{Static Listening}

Overall, At-Risk individuals demonstrated lower \textit{Compressiblity} as a group than No-Risk, the difference, albeit insignificant. However, this trend was observed consistently across the entire listening and for all the top \textit{n} tracks. This lack of difference observed between the groups can be attributed to two reasons. Firstly, though individuals At-Risk are characterised by their preference for music tagged as \textit{Sad}, they are also characterised by a tendency to gravitate towards genres such as \textit{neo-psychedilic-dream-pop} and \textit{Indie-Alternative-pop} genres. The very nature of the latter genre indeed may offset the overall lyrical compressiblity scores as they are inherently repetitive and hence highly compressible \cite{markus2019}. Second, the relation between \textit{Compressibility} and the absolute information content, which can often contradict each other. A higher compression score in a shorter song may be equivalent to a longer song with a lower compression score in terms of the information content. Hence, the AIC metric may be a more apt measure as it can capture the size of a song in a more perceptually relevant fashion when compared to \textit{Compressibility}. In line with our hypothesis, the AIC metric shows a significant difference consistently across top tracks (Table \ref{tab:aic_table}, Fig. \ref{fig:info_all}), suggesting that the At-Risk group prefers songs with greater information content than the No-Risk group. Furthermore, this effect was observed across the entire listening history and top \textit{n} (100, 250, 500) tracks thereby adding to the consistency of the results.


The positive correlation between Valence and \textit{Compressibility} suggests that \textit{Happy} music indeed has higher lyrical simplicity than \textit{Sad} music, which confirms similar observations from prior work \cite{brattico2011} \cite{fiveash2016}. The quadrant specific analysis also provides further insight into group preferences within certain emotion categories. The \textit{Sadness} quadrant has a lower median for \textit{Compressibility} for the At-Risk group, suggesting a preference for songs with greater lyrical complexity. This is also supported by group differences observed in AIC, with At-Risk showing a higher median for tracks belonging to emotion quadrants representing \textit{Happiness}, \textit{Sadness}, and \textit{Tenderness}. This may be indicative of general listening preferences, which is also reflected when considering the entire 6-month listening history and top $n$ tracks.

The significance of age, in relation to AIC, is suggestive of evolving lyrical preferences, as in \cite{varnum2021}. While mean AIC for At-Risk is comparable for both age groups, No-Risk youth has a higher AIC. This is possibly due to popular music trends where high AIC genres such as Hip-Hop/Rap have become dominant \cite{fossi2021}. Considering age alongside cultural factors in the future may reveal more insights, especially as music tastes are developed during youth \cite{holbrook89}, suggesting that music should not only be studied in relation to other users but also its own evolution.

The notion of compressibility is also tied to the idea of cognitive load. Songs with higher \textit{compressibility} have highly repetitive predictable patterns, and hence likely low cognitive load. If we consider low lyrical simplicity songs, representative of higher cognitive load, this could also explain At-Risk preference music with greater lyrical content. The tendency of At-Risk individuals with such music (i.e., low \textit{Compressiblity} and high AIC) suggests that they may use such music to distract themselves from a reality they perceive to be adverse, indicative of avoidant coping \cite{Miranda2009MusicLC}. This result is very much in line with studies that show higher load activities act as a distraction \cite{Hu2021CanIC} \cite{Stewart2019} from negative thoughts for At-Risk individuals.

One caveat is that the division of tracks into quadrants was based on Valence and Energy derived from acoustic content instead of lyrics. However, it is common for music and lyrics to be congruent in terms of their respective emotional connotations, though joint usage of acoustic and lyrical features may yield better results \cite{mihalcea2012lyrics}. Future work can look at the interplay between emotional connotations of lyrics and lyrical complexity along with other linguistic features. While we collect lyrics for the majority of the dataset, it is largely limited to English music with lyrics. Further, future work can consider the role of instrumentals in At-Risk listening; it can also explore lyrical simplicity preferences in other languages and investigate the role of cultural influences, though we do not expect much difference with the former as our approach is language agnostic.  

\subsection{Dynamic Listening}

As hypothesised, At-Risk individuals demonstrated higher inter-session AIC variability. On the other hand, the higher intra-session AIC variability for At-Risk contradicts our hypothesis that At-Risk individuals prefer a certain level of information content during a local period (i.e. session) of continuous listening. This result deserves further exploration as there may be other confounds to consider. Future work could explore intra-session variability depending on the predominant emotion.


The lack of group differences in \textit{Compressiblity} can be attributed to the nature of the metric itself, which ranges from 0-1, with most songs clustered around a value of 0.4/0.5. Hence, the difference in standard deviation values between the groups is likely to be marginal or insignificant when compared to AIC, which has a larger range. 

In summary, we demonstrate associations between lyrical attributes and online music consumption of individuals at risk for depression. Our results validate our hypothesis that the At-Risk group prefers songs with greater information content, especially for songs characterised as \textit{Sad}. Furthermore, we found that At-Risk individuals also have greater variability of the Absolute Information Content across their listening history. Finally, these results also motivates the use of lyrical features, with acoustics and tag-based features, in building a multi-modal system to predict risk for depression. While such analysis stops short of providing clinical solutions for depression, this information can be used for the early identification of depression symptoms and even personalised recommender systems.

\section{Acknowledgements}

We thank Shivani Hanji for their valuable input and assistance.